\documentclass[journal]{IEEEtran}
\usepackage[comma,numbers,square,sort&compress]{natbib}
\usepackage{subfigure}
\usepackage{color}
\usepackage{colortbl}
\usepackage{graphicx}
\usepackage{float}
\usepackage{amsmath,latexsym,amssymb,amsthm,array,amsfonts,algorithm,algpseudocode,booktabs,graphicx,subfigure,multirow,cuted,stfloats}
\usepackage[sort&compress]{natbib}
\usepackage{url}
\usepackage{multirow}
\usepackage{lscape}
\usepackage{threeparttable}
\usepackage{nomencl}
\makenomenclature
\usepackage{makecell}
\usepackage{float}
\usepackage{enumerate}
\usepackage{bm}
\definecolor{mygray}{gray}{0.8}
\theoremstyle{definition}
\usepackage{array}

\algrenewcommand{\algorithmiccomment}[1]{\hskip1em$/*$ #1 $*/$}

\newcommand{\tabincell}[2]{\begin{tabular}{@{}#1@{}}#2\end{tabular}}
\newlength\savewidth
\newcommand\shline{\noalign{\global\savewidth\arrayrulewidth
                            \global\arrayrulewidth 1.2pt}%
                   \hline
                   \noalign{\global\arrayrulewidth\savewidth}}

\newcolumntype{I}{!{\vrule width 1.2pt}}
\newlength\savedwidth

 \newtheorem{definition}{Definition}
 \newtheorem{remark}{Remark}

\begin{document}
\hyphenpenalty=5000
\tolerance=1200
%
\title{A Novel Dual-Stage Evolutionary Algorithm for Finding Robust Solutions}

\author{Wei Du, \emph{Member, IEEE}, Wenxuan Fang,
        Chen Liang,
        Yang Tang, \emph{Fellow, IEEE}, \\
        and Yaochu Jin, \emph{Fellow, IEEE}\\
\thanks{This work was supported by the National Natural Science Foundation of China (Key Program: 62136003), National Natural Science Foundation of China (62173144, 62373154), Shanghai Rising-Star Program (22QA1402400) and the Programme of Introducing Talents of Discipline to Universities (the 111 Project) under Grant B17017. }

\thanks{W. Du, W. Fang, C. Liang, and Y. Tang are with the Key Laboratory of Smart Manufacturing in Energy Chemical Process, Ministry of Education, East China University of Science and Technology, Shanghai 200237, China (e-mail: duwei0203@gmail.com; y30210943@mail.ecust.edu.cn; chenliang@ecust.edu.cn; tangtany@gmail.com, yangtang@ecust.edu.cn).}

\thanks{Y. Jin is with the School of Engineering, Westlake University, Hangzhou 310030, China (e-mail: jinyaochu@westlake.edu.cn)}
}

\markboth{}%
{Shell \MakeLowercase{\textit{et al.}}: Bare Demo of IEEEtran.cls
for Journals}

\maketitle

\begin{abstract}
In robust optimization problems, the magnitude of perturbations is relatively small.
Consequently, solutions within certain regions are less likely to represent the robust optima when perturbations are introduced.
Hence, a more efficient search process would benefit from increased opportunities to explore promising regions where global optima or good local optima are situated.
In this paper, we introduce a novel robust evolutionary algorithm named the dual-stage robust evolutionary algorithm (DREA) aimed at discovering robust solutions.
DREA operates in two stages: the peak-detection stage and the robust solution-searching stage.
The primary objective of the peak-detection stage is to identify peaks in the fitness landscape of the original optimization problem.
Conversely, the robust solution-searching stage focuses on swiftly identifying the robust optimal solution using information obtained from the peaks discovered in the initial stage.
These two stages collectively enable the proposed DREA to efficiently obtain the robust optimal solution for the optimization problem.
This approach achieves a balance between solution optimality and robustness by separating the search processes for optimal and robust optimal solutions.
Experimental results demonstrate that DREA significantly outperforms five state-of-the-art algorithms across 18 test problems characterized by diverse complexities.
Moreover, when evaluated on higher-dimensional robust optimization problems (100-$D$ and 200-$D$), DREA also demonstrates superior performance compared to all five counterpart algorithms.
\end{abstract}

\begin{IEEEkeywords}
Evolutionary robust optimization, evolutionary algorithm, dual-stage strategy, peak detection
\end{IEEEkeywords}

%
\IEEEpeerreviewmaketitle

\section{Introduction}
Robust optimization is a research field that deals with the optimization problem in which the uncertainties comes from decision variables or environmental parameters \cite{jin2005evolutionary,beyer2007robust}.
Solutions that are comparatively less sensitive to small perturbations are called robust solutions.
From a practical standpoint, users may be more interested in searching for robust solutions because the perturbations often cannot be avoided in practice \cite{du2018robust,fan2008improved,du2022searching}.
A number of effective approaches have been applied to solve robust optimization problems, including Taguchi method in the early days, mathematical programming approach and evolutionary algorithms (EAs) \cite{beyer2007robust}.
Due to the major advantages of EAs that can be applied on a wide range of problems without much problem specific knowledge, EAs have received increasing attention in the past two decades.
Then a new research field called evolutionary robust optimization comes up.

Generally, three main issues should be well addressed in evolutionary robust optimization: 1) how to measure robustness of a candidate solution, 2) how to quantify robustness, and 3) how to find the robust optimal solution effectively.

First, many attempts have been made on evaluating the robustness degree of a candidate solution. The commonly used ones are statistical measures, which include expectation, variance or the integration of them \cite{beyer2007robust,lei2013robust,sun2011crashworthiness}.
For robust multi-objective optimization problems, a number of robustness measures have been specifically designed, such as dominance robustness, preference robustness and $r$-dominance \cite{bui2010robustness,li2015new}.
Second, when the robustness measure is determined, one should consider how to quantify it in the context of evolutionary robust optimization.
Sampling is the most widely used technique in practice.
The simplest sampling technique is Monte-Carlo sampling which estimates the robustness measure of one solution by averaging over a number of it neighboring points \cite{wiesmann1998robust}.
The main drawback of Monte-Carlo sampling is computationally inefficient.
To overcome this drawback, one way is to use tailored sampling Monte-Carlo sampling techniques.
The representative methods are Latin hypercube sampling, importance sampling and one sample strategy (also known as implicit averaging) \cite{liu2008monte,fei2018new,tsutsui1997genetic}.
Recently, a novel robust EA called robust particle swarm optimizer with implicit averaging and condition (RPSO\_IC) is proposed, which develops a constraint to improve the reliability of implicit averaging-based methods \cite{mirjalili2019improving}.
In RPSO\_IC, the number of previously sampled points in the neighbourhood of a solution is counted and used to check how reliable the robustness measure of a solution is while searching for the robust optimum solution.
The other way is to use metamodel approach to save computational resources.
The successful applications in evolutionary robust optimization include Kriging metamodels, radial basis function-based model, and interpolation/regression models, among others \cite{kruisselbrink2010robust,liu2016surrogate,ong2006max,paenke2006efficient}.

After the robustness measure and the quantification method are determined, the following most important issue is to develop optimization techniques for searching robust optimal solutions effectively.
A useful idea is to change the structure of the problem by adding additional objective or constraint in the original robust optimization problem.
For example, optimizing the robustness measure (e.g., the expectation or the variance) can be used as an extra objective function \cite{ray2002constrained,jin2003trade}.
Then the original single-objective optimization problem is converted to a multi-objective optimization one.
Another example is that the robust constraints are considered in the original optimization problem, then constrained EAs are adopted for the efficient search of robust solutions \cite{deb2006introducing}.
For robust multi-objective optimization problems, there have been a number of studies proposed in recent years.
For example, the coevolutionary framework is introduced to solve the robust multi-objective optimization problem \cite{meneghini2016competitive}.
Instead of incorporating robustness measures in the objective function or in the constraints, two populations representing candidate solutions and perturbations respectively are utilized to compete for identifying the robust solutions.
A novel EA called multi-objective optimization EA with robustness enhancement is developed by maintaining an external archive, from which the robustness is enhanced by deleting existing solutions with bad optimal performance under the current environment \cite{he2019evolutionary}.
Another example for solving robust multi-objective optimization problems is a novel robust multi-objective optimization EA (RMOEA) \cite{he2018robust}.
RMOEA first performs the multi-objective optimization, and then sequently conduct performance estimation of optimal solutions under disturbance, archive update, robust region detection, and selection of robust solutions to construct the robust optimal front.
For other recent advances, the reader is referred to \cite{chan2010evolutionary,asafuddoula2014six,chen2022adaptive} and references therein.

Although a number of useful techniques have been developed for searching robust optimal solutions, the problem's features induced by introducing perturbations can be better utilized.
For robust optimization problems, the amount of perturbation is relatively small, which means points in certain regions are less likely to become the robust optima when the perturbation is imposed.
These regions include the ones with poor fitness or flat regions according to intuitive judgement.
In other words, the robust optimal solutions are more likely to evolve from the points with good fitness values of the original problem without considering the uncertainties.
Therefore, the search will be more efficient if we have more opportunities to search around promising regions (e.g., global optima or good local optima).

Motivated by the above discussions, in this article, we propose a novel dual-stage robust evolutionary algorithm, termed DREA, designed specifically for addressing robust optimization problems.
DREA operates through two distinct stages: the peak-detection stage and the robust solution-searching stage.
During the peak-detection stage, we identify the peaks (i.e., the local maxima within the original fitness landscape aimed at maximizing the problem) of the original optimization problem with an external archive which stores the historical population and fitness values.
Subsequently, in the robust solution-searching stage, we use the information of the peaks obtained from the peak-detection stage to quickly locate the robust optimal solution.
The proposed DREA can efficiently obtain the robust optimal solution of the optimization problem.
It should be mentioned that unlike other methods for evolutionary robust optimization, this work uniquely utilizes the features of the problems that are induced by the introduction of perturbations to facilitate the search process.
These features highlight that points in specific regions, such as those with poor fitness or flat regions, are less likely to represent the robust optima because the amount of perturbation is often relatively small.
Consequently, we employ points with favorable fitness values from the original problem (disregarding uncertainties) to assist in the search for robust optimal solutions.
The contributions of this article can be summarized as follows.
\begin{itemize}
  \item [1)]
  This work proposes a novel dual-stage evolutionary algorithm called DREA for robust optimization problems. DREA consists of two stages: the peak-detection stage and the robust solution-searching stage. DREA utilizes the global optimal or good local optimal solutions of the original optimization problem (i.e., without considering the perturbations) to expedite the location of the robust optimal solution. To the best of our knowledge, this is the first attempt to use solutions from the original optimization problem to facilitate the search for the robust optimal solution. The dual-stage framework is also quite novel for evolutionary robust optimization. As a result, a seamless integration of robustness and optimality is achieved.
  \item [2)]
  In the first stage of DREA, a specified number of peaks are located by using an external archive that stores historical individuals and their fitness values during the search.
  While other peak detection methods have been proposed for solving multimodal optimization problems, such as the adaptive peak detection (APD) method proposed to identify potential optimal solutions \cite{cheng2017evolutionary}, the objective in the peak-detection stage of DREA, as proposed in this article, is to detect a specific number of peaks rather than finding the optimal solutions.
  This distinction sets it apart from both APD and other peak detection methods for multimodal optimization problems.
  In the second stage of DREA, the peaks (with good objective values) found in the first stage are used to guide the population to search around the promising regions where the points are more likely to evolve into robust optimal solutions.
  This operation significantly reduces the time taken to find the robust optimal solution.
\end{itemize}

The remainder of this article is organized as follows. Section II provides the background information.
In Section III, the proposed DREA is introduced in detail.
Section IV gives a series of experimental studies to validate the effectiveness of DREA.
Finally, concluding remarks are provided in Section V.

\section{Background}
In this section, we will begin by providing background information on evolutionary robust optimization.
We will then introduce recent research on dual-stage EAs.
Finally, we will present information on peak detection methods proposed in recent years.

\subsection{Evolutionary robust optimization}
A robust optimization problem can be defined as follows (for the maximization of a problem):
\begin{align}
\left.
\begin{array}{ll}
\textrm{maximize}&\hspace{0.28cm} f(\textbf{x},\bm{\delta}) \\
\textrm{s.t.}&\hspace{0.25cm} \textbf{x}\in\Omega\\
\end{array}
\right.
\label{eqproblem1}
\end{align}
where $f(\cdot)$ is objective function of the original optimization problem, $\textbf{x}=[x_1,x_2,...,x_D]^T$ is the decision vector, $\Omega$ denotes the feasible decision space, $\bm{\delta}$ is the perturbation imposed on the decision vector.

A solution $\textbf{x}^{\ast}$ is called the robust optimal solution if $\textbf{x}^{\ast}$ is the global maximum of $f(\textbf{x},\bm{\delta})$.
To obtain $\textbf{x}^{\ast}$, we often optimize the mean effective objective function of the original optimization problem, which is defined as follow:
\begin{equation}
f^{\textrm{eff}}(\textbf{x})=\frac{1}{|\mathcal{B}_{\bm{\delta}}(\textbf{x})|}\int_{\textbf{y}\in\mathcal{B}_\delta(\textbf{x})}f(\textbf{y})d\textbf{y},
\label{eqprofi}
\end{equation}
where $\mathcal{B}_{\bm{\delta}}(\textbf{x})$ denotes a $\bm{\delta}$-neighborhood of $\textbf{x}$, $|\mathcal{B}_{\bm{\delta}}(\textbf{x})|$ indicates the related hypervolume of the neighborhood.
It is shown that $f^{\textrm{eff}}(\textbf{x})$ can be calculated by averaging the objective values of the solutions in $\mathcal{B}_{\bm{\delta}}(\textbf{x})$.
In evolutionary robust optimization, EAs are used as the optimizer for optimizing $f^{\textrm{eff}}(\textbf{x})$.

\subsection{Dual-stage EAs}
Dual-stage or two-stage EAs divide the search into two phases.
In each stage, the purpose of the population is different, which aims to efficiently find the target solutions.
Under the framework of EA, dual-stage-based approaches have been successfully adopted to solve different kinds of optimization problems.

For many-objective optimization problems, the dual-stage strategies are introduced to maintain the population with both promising convergence and diversity \cite{hu2016many,zhu2016two,sun2018new,ming2022two}.
The first stage guarantees the convergence of the solutions, while the second stage promotes the diversity.

For constrained optimization problems, the dual-stage strategies are proposed to facilitate the population to efficiently search for the feasible solutions by utilizing both feasible and infeasible solutions \cite{fan2019push,ming2021novel,ming2021simple}.
In the first stage, the constraints of the problem are often ignored, the purpose of which is to explore the entire search space regardless of feasibility.
In the second stage, the population focuses on the exploitation of the feasible region, which considers all the constraints of the problem.

In this work, a dual-stage EA is proposed for solving robust optimization problems.
The dual-stage strategy can better utilize the solutions of both the original optimization problem and the robust optimization problem.

\subsection{Peak detection}
Peak detection is a method used to identify the global maxima (i.e., peaks) in the fitness landscape of an optimization problem.
There are a number of peak detection methods that have been proposed, especially for solving multimodal optimization problems.
For instance, a binary cutting-based adaptive peak detection (APD) method was proposed to identify peaks where optimal solutions may exist \cite{cheng2017evolutionary}.
The detection is based on the approximated fitness landscape generated beforehand.
The purpose of this method is to identify all potential peaks where optimal solutions may exist.
Once the peaks are detected, a local search is performed inside each of them.
Other methods utilize either an individual or the population to locate global optima \cite{chen2019distributed,gao2013cluster}, which can also be considered a form of niching scheme.
For example, the method proposed in \cite{chen2019distributed} enables each individual to act as a distributed unit for tracking a peak.
In \cite{gao2013cluster}, the entire population is divided into several subpopulations, each maintained for different peaks.
However, in this article's peak detection method, our objective is to detect the designated number of peaks, rather than finding global optimal solutions. This makes our approach distinct from both APD and other peak detection methods for multimodal optimization problems.

It is worth noting that the peak detection discussed in this article falls under fitness landscape analysis (FLA), an important topic in evolutionary computation in the last three decades \cite{prugel2011maximum}.
The goal of FLA is to identify features of the landscape for a given optimization problem, aiding in a better understanding of how and where EAs operate.
Widely-used FLA methods include fitness distance correlation, exploratory landscape analysis (ELA), and local optima networks (LONs), among others.
Fitness distance correlation is designed to provide a global view of the fitness landscape by considering the correlation between fitness values and their distances from the global optimum \cite{jones1995fitness}.
ELA aims to extract high- and low-level features of fitness landscapes, facilitating the classification of optimization problems into several categories and paving the way for automatic algorithm selection \cite{mersmann2011exploratory,munoz2014exploratory}.
LONs serve as a coarse-grained model for discrete (combinatorial) fitness landscapes and have been further extended to continuous fitness landscapes. In this model, nodes represent local optima, and edges indicate search transitions based on an exploration search operator \cite{ochoa2008study,adair2019local}.
However, the main distinction between the peak detection method discussed in this article and the aforementioned FLA methods lies in the focus on local features (i.e., peaks) of the fitness landscape rather than the global structure of the search space.

\section{The Proposed DREA}
\subsection{Outline of DREA}
\begin{figure}[htbp]
\centering
\includegraphics[width=9cm]{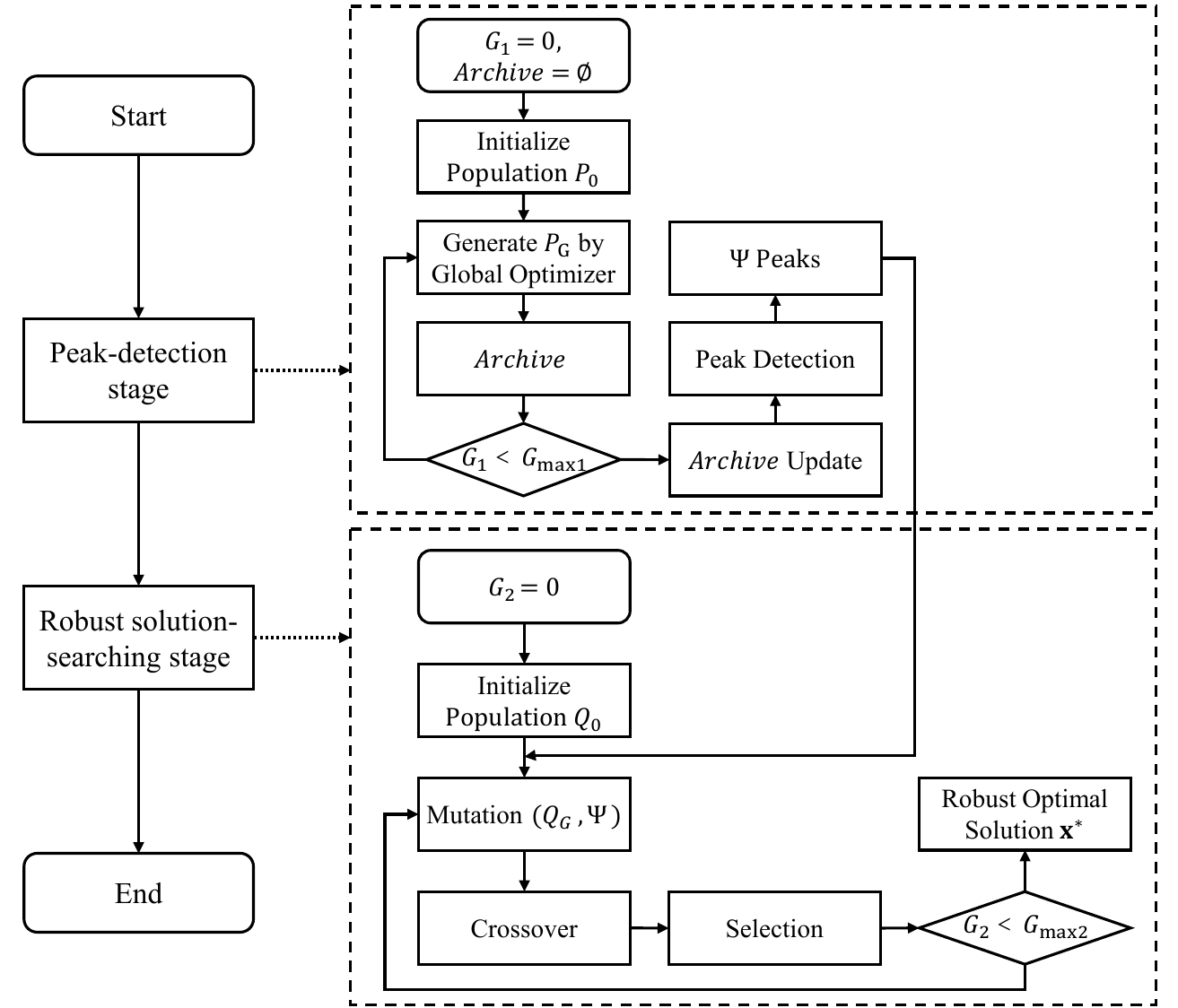}\\
\caption{The overall flowchart of DREA.}\label{flowchart}
\end{figure}

As previously mentioned, the proposed DREA consists of two stages: the peak-detection stage and the robust solution-searching stage.
DREA starts with the peak-detection stage, which identifies the peaks of the original optimization problem.
These identified peaks serves as guides for the subsequent robust solution-searching stage, where the final output of DREA is determined.
The overall flowchart of DREA is depicted in Fig. \ref{flowchart}.
The process begins with the peak-detection stage illustrated in Fig. \ref{flowchart}.
Here, an external $Archive$ is initialized to store historical individuals and their fitness values across a predefined number of generations denoted as $G_{max1}$.
Subsequently, the updated $Archive$ is employed to execute peak detection, yielding $\Psi$ peaks.
These identified $\Psi$ peaks are utilized to guide the population in the mutation process within the robust-solution searching stage, ultimately leading to the determination of the robust optimal solution after a predefined number of generations denoted as $G_{max2}$.

The outline of DREA is provided in Algorithm \ref{alg_overall_framework}.
In Algorithm \ref{alg_overall_framework}, Lines 2-7 describe the peak-detection stage.
First, we initialize a population $P_0$, the size of which is $N$ (Line 2).
Then we use a global optimizer to optimize the original optimization problem without considering the perturbation $\bm{\delta}$ (Line 3).
During the search, all the historical individuals and their fitness values are stored in an $Archive$, the element of which is represented by a pair $(\textbf{x}, f(\textbf{x}))$.
Then the size of $Archive$ is checked if it is larger than the predefined value (Lines 4-6).
The $Archive$ is used to detect the peaks of the original optimization problem (Line 7).
Lines 9-10 show the robust solution-searching stage.
Based on the peaks obtained from the first stage, we conduct the search for the robust optimal solution of the problem by a global optimizer (Line 10).
The details of the two stages are introduced in the following subsections.

{\linespread{1}
\begin{algorithm}[!ht]
\small
\caption{Procedure of DREA}
\label{alg_overall_framework}
\hspace*{0.02in} {\bf Input:}  $N$ -- the population size \\
\hspace*{0.4in} $\bm{\delta}$ -- the perturbation imposed on the decision vector \\
\hspace*{0.38in} $N_a$ -- the archive size \\
\hspace*{0.38in} $N_p$ -- the number of the peaks to detect
\begin{algorithmic}[1]
\Ensure{$\textbf{x}^*$ -- the robust optimal solution}
\State /****** Peak-detection stage ******/
\State $P_{0}=\{\textbf{x}_{1,0},\textbf{x}_{2,0},...,\textbf{x}_{N,0}\}$  \Comment{Initialize the population in the peak-detection stage}
\State $Archive = \textit{Global\_Optimizer}(P_{0})$ \Comment{$Archive$ stores all the historical population and fitness values during the search for the original optimization problem (the element in $Archive$ is represented by a pair $(\textbf{x}, f(\textbf{x}))$)}
\Comment{Algorithm \ref{global_optimizer_1}}
\If	{$|Archive|>N_a$}
    \State $Archive = \textit{Archive\_Update}(Archive, N_a)$
    \Comment{Algorithm \ref{archive_update}}
\EndIf
\State    $\Psi \leftarrow \textit{Peak\_Detection} (Archive, N_{p})$
\State /****** Robust solution-searching stage ******/
\State $Q_{0}=\{\textbf{x}_{1,0},\textbf{x}_{2,0},...,\textbf{x}_{N,0}\}$  \Comment{Initialize the population in the robust solution-searching stage}
\State $\textbf{x}^* = \textit{Global\_Optimizer}(Q_{0},\bm{\delta},\Psi)$
\end{algorithmic}
\end{algorithm}
}

\subsection{Peak-detection stage}\label{seciiib}
The main purpose of the peak-detection stage is to detect the peaks of the fitness landscape of the original optimization problem.
The perturbations are not considered in this stage.
The peaks are detected by means of an $Archive$ that stores the historical individuals and their fitness values during the search for the original optimization problem.
The element in $Archive$ is represented by a pair $(\textbf{x}, f(\textbf{x}))$.
The details of constructing $Archive$ are shown in Algorithms \ref{global_optimizer_1}-\ref{archive_update}.

In Algorithm \ref{global_optimizer_1}, a global optimizer is utilized to search for the optimal solution of $f(\textbf{x})$ with a initialized population $P_0$.
The search is performed within fixed iterations.
The individuals generated during the search are saved in $Archive$ (Lines 3 and 6).

If the size of $Archive$ exceeds $N_a$, we will remove the additional points in Algorithm \ref{archive_update}.
The update procedure is divided into two steps.
The first step is to keep the point with largest $f(\textbf{x})$ value in $Archive$ (Line 1).
The purpose is to avoid discarding the best solution obtained by the global optimizer.
Then the rest $N_a-1$ points are randomly selected from the remaining points, the purpose of which is to fairly depict the fitness landscape (Lines 2-3).

{\linespread{1}
\begin{algorithm}[!h]
\small
\caption{$\textit{Global\_Optimizer}(P_0)$}
\label{global_optimizer_1}
\hspace*{0.02in} {\bf Input:} $P_0$ -- the initialized population which is represented by $\{\textbf{x}_{1,0},\textbf{x}_{2,0},...,\textbf{x}_{N,0}\}$ \\
\hspace*{0.02in} {\bf Output:}  $Archive$ -- the archive stores all the historical population and fitness values during the search for the original optimization problem
\begin{algorithmic}[1]
\State $Archive = \emptyset$
\Comment{Initialize the archive}
\State $G = 0$
\Comment{Initialize the generation counter}
\State $Archive = Archive \cup \{(\textbf{x}_{1,G},f(\textbf{x}_{1,G})),(\textbf{x}_{2,G},f(\textbf{x}_{2,G})),
...,$ $(\textbf{x}_{N,G},f(\textbf{x}_{N,G}))\}$
\While  {$G < G_{max1}$}
    \State Generate the offspring $P_{G+1}$ from $P_G$
    \State $Archive = Archive \cup \{(\textbf{x}_{1,G+1},f(\textbf{x}_{1,G+1})),(\textbf{x}_{2,G+1},
$ $f(\textbf{x}_{2,G+1})),...,(\textbf{x}_{N,G+1},f(\textbf{x}_{N,G+1}))\}$
    \State $G=G+1$
\EndWhile
\end{algorithmic}
\end{algorithm}
}

{\linespread{1}
\begin{algorithm}[!h]
\small
\caption{$\textit{Archive\_Update}(Archive, N_a)$}
\label{archive_update}
\hspace*{0.02in} {\bf Input:}  $Archive$ -- the archive stores all the historical population and fitness values during the search for the original optimization problem \\
\hspace*{0.38in} $N_a$ -- the archive size \\
\hspace*{0.02in} {\bf Output:}  $Archive$ -- the archive with the predefined size
\begin{algorithmic}[1]
\State  $(\textbf{x}_{i},f(\textbf{x}_{i})) = \max\limits_{f(\textbf{x})}(Archive)$  \Comment{Select the point with the largest $f(\textbf{x})$ value in $Archive$}
\State  $Archive\_{temp} =  Archive  \backslash  (\textbf{x}_{i},f(\textbf{x}_{i}))$   \Comment{Remove the point that has been selected}
\State Randomly select $N_a-1$ points $\{(\textbf{x}_{1},f(\textbf{x}_{1}),(\textbf{x}_{2},f(\textbf{x}_{2})),$
$...,(\textbf{x}_{N_a-1},f(\textbf{x}_{N_a-1}))\}$ from $Archive\_temp$
\State $Archive = \emptyset \cup (\textbf{x}_{i},f(\textbf{x}_{i})) \cup \{(\textbf{x}_{1},f(\textbf{x}_{1}),(\textbf{x}_{2},f(\textbf{x}_{2})),$
$...,(\textbf{x}_{N_a-1},f(\textbf{x}_{N_a-1}))\}$
\end{algorithmic}
\end{algorithm}
}

When $Archive$ is obtained, we will detect the specified number of the peaks.
The procedure of peak detection is provided in Algorithm \ref{alg_peak_detection}.
For each point in $Archive$, we need to check whether it is a peak.
The main idea of Algorithm \ref{alg_peak_detection} is to detect the peaks according to the fitness value f(\textbf{x}) of each point in $Archive$.
The fitness values are sorted in descending order.
Before we conduct the peak detection, we provide several definitions.

\begin{definition}\label{def_peakset}
\emph{(Peak Set)}\\
A peak set is defined as the point set that contains a peak and its neighbouring points:
\begin{equation}
 PeakSet_k=\{Point^k_i \mid i=1, ..., p\},
\end{equation}
where $p$ is the total number of the points in $PeakSet_k$.
\end{definition}

\begin{definition}\label{def_peak}
\emph{(Peak)}\\
A point $Point_i$ is defined as a peak if it has the maximum fitness value in the $PeakSet$ it belongs to:
\begin{equation}\label{eq_peak}
 Peak_k = \max \limits_{f(\textbf{x})}\{ Point^k_i \mid Point^k_i \in PeakSet_k, i=1,...,p \}
\end{equation}
where $p$ is the total number of the points in $PeakSet_k$.
\end{definition}

\begin{definition}\label{def_dist}
\emph{(Distance between a point and a peak set)}\\
For a point, the distance between this point and a peak set is defined as follows:
\begin{multline}\label{eq_dist_point2peak}
  Dist(Point_j, PeakSet_k) \\
  = \min \parallel Point_j - Point^k_{i} \parallel_2, i=1, \ldots, p,
\end{multline}
where $Point_j$ is any single point, $Point^k_{i}$ is the point in $PeakSet_k$, $\parallel \cdot  \parallel_2$ denotes the Euclidean distance.
\end{definition}
From Definition \ref{def_dist}, it implies that for a point that does not belong to $PeakSet_k$, $Dist(Point_j, PeakSet_k)$ is defined as the minimum distance between $Point_j$ and all the points in $PeakSet_k$; while for point that belongs to $PeakSet_k$, $Dist(Point_j, PeakSet_k)=0$.

\begin{definition}\label{def_neigh}
\emph{(Neighbourhood of a point with regard to a peak set)}\\
For a point, the neighbourhood of this point with regard to a peak set is defined as follows:
\begin{multline}\label{eq_dist_point2peak}
  \mathcal{B}_{\theta,r}(Point_j) \\
  = \{ c \mid \angle(proj_\textbf{x}(\overrightarrow{ab}), proj_\textbf{x}(\overrightarrow{ac})) \leq \theta, |proj_\textbf{x}(\overrightarrow{ac})| \leq r \},
\end{multline}
where $Point_j$ is any single point, $a$ is used to represent $Point_j$, $b$ denotes the peak (i.e., $Peak_k$) of a peak set (i.e., $PeakSet_k$), $c$ indicates any point in the neighbourhood of $Point_j$ with regard to $PeakSet_k$, $\theta$ is the angle predefined; $proj_\textbf{x}(\cdot)$ denotes the projection of the vector in $\textbf{x}$-plane; $\angle(\cdot,\cdot)$ represents the angle between the projections of two vectors, $r= Dist(Point_j, PeakSet_k)$.
\end{definition}
From Definition \ref{def_neigh}, a point $c$ is in $\mathcal{B}_{\theta,r}(Point_i)$ if the projection of $\overrightarrow{ac}$ satisfies both the angle and the length requirements.

\begin{remark}
It is observed that the neighbourhood of a point with regard to a peak is defined by an angle $\theta$ and a distance $r$.
The distance determines how far the neighbourhood can reach, while the angle defines the range of the neighbourhood.
Together, the distance and the angle define a sector-shaped region in the projection plane, within which the points can be regarded as belonging to the neighbourhood of a specific point.
\end{remark}

\begin{figure}[htbp]
\centering
\includegraphics[width=8cm]{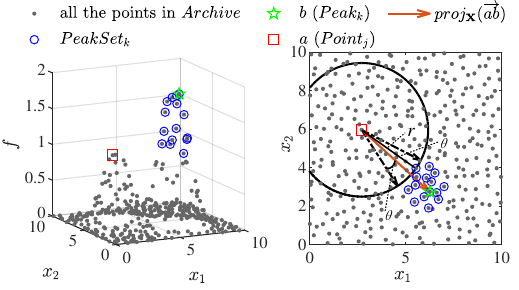}\\
\caption{The illustration of the neighborhood of $Point_j$ with regard to $PeakSet_k$. (a) The $(\textbf{x},f)$-plot; (b) The projection in $\textbf{x}$-plane. }\label{fig_theta_neighbor_2d}
\end{figure}

We provide a 2-$D$ example in Fig. \ref{fig_theta_neighbor_2d} to illustrate the above definitions.
Fig. \ref{fig_theta_neighbor_2d}(a) plots the archive points in 3-$D$ space; Fig. \ref{fig_theta_neighbor_2d}(b) plots the points in $\textbf{x}$-plane from the top view.
In Fig. \ref{fig_theta_neighbor_2d}, the gray dots represent all the points in $Archive$.
The blue circles compose the peak set (i.e., $PeakSet_k$, Definition \ref{def_peakset}).
The green pentagram is the peak (i.e., $Peak_k$, Definition \ref{def_peak}) of $PeakSet_k$.
The red square indicates the point (i.e., $Point_j$) the neighbourhood of which with regard to $PeakSet_k$ we want to check.
The red arrow shows the distance between a point and a peak set (i.e., $Dist(Point_j, PeakSet_k)$, Definition \ref{def_dist}).
The region of a circular sector that is defined by the angle $\theta$ and the radius $r$ represents the neighbourhood of $Point_j$ with regard to $PeakSet_k$ (i.e., $\mathcal{B}_{\theta,r}(Point_j)$, Definition \ref{def_neigh}).

According to the above definitions, the peak detection is conducted based on the following procedure.
We select the point in $Archive$ according to the fitness value f(\textbf{x}) in descending order.
For each point $Point_j$ selected, if there are other smaller points in the neighbourhood of $Point_j$ with regard to $PeakSet_k$ (i.e., $\mathcal{B}_{\theta,r}(Point_j)$), $Point_j$ is regarded as a new peak; otherwise, $Point_j$ belongs to $PeakSet_k$.
For the example in Fig. \ref{fig_theta_neighbor_2d}, $Point_j$ is a new peak.

The pseudocode of the peak detection procedure is provided in Algorithm \ref{alg_peak_detection}.
Lines 1-2 predefine the value of $\theta$ in Eq. (\ref{eq_dist_point2peak}) and the number of the peaks to detect.
Lines 3-23 show the peak detection procedure.
Each point in $Archive$ is checked whether it belongs to an existing peak set (Lines 7-11) or it is a peak of a new peak set (Lines 13-15).
Algorithm \ref{alg_peak_detection} returns the required number of the peaks detected (Lines 22-23).

It it worth mentioning that the peak-detection stage reminds us of the refinement procedure for initial solutions in EAs.
The purpose of population refinement in the initialization stage is to increase the likelihood of finding good solutions within the vast search space.
In recent years, variants of refinement methods for initial solutions have been proposed, including opposition-based learning (OBL)-based initialization \cite{rahnamayan2008opposition}, initialization based on mirror partition \cite{yao2023solution}, and initialization using heuristics \cite{azad2017heuristic}, among others.
Specifically, OBL-based initialization generates the opposite point of each solution in the initial population, from which the fitter one is selected as the initialization population.
In population initialization based on mirror partition, the decision space is divided into several subdomains. The sampling probability is then updated based on the performance of individuals among different subdomains, and this process is repeated until a high-quality population is obtained.
In initialization using heuristics, a well-designed heuristic provides good initial solutions in terms of both cost and feasibility, aiming to reduce evolution time.
However, strictly speaking, the peak-detection stage proposed in this article does not belong to a population refinement method.
The main reasons are as follows:
First, the peak-detection stage only identifies a number of peaks in the provided optimization problem, rather than generating an initial population.
Second, the peaks obtained in the peak-detection stage serve as guidance for the subsequent global search, rather than acting as the starting points of the following search.

{\linespread{1}
\begin{algorithm}[!h]
\small
\caption{$\textit{Peak\_Detection}(Archive, N_p)$}
\label{alg_peak_detection}
\hspace*{0.02in} {\bf Input:} $Archive$ -- the archive stores all the historical population and fitness values during the search for the original optimization problem \\
\hspace*{0.36in} $N_{p}$ -- the number of the peaks to detect\\
\hspace*{0.02in} {\bf Output:}  $\Psi = \{Peak_k \mid k=1,...,n\}$ -- the peaks detected
\begin{algorithmic}[1]
\State  $\theta = \pi/12$ \Comment{Predefine the value of $\theta$ in Eq. (\ref{eq_dist_point2peak})}
\State  $n = 1$ \Comment{Initialize the number of the peaks detected}
\State  $Point_j = \max\limits_{f(\textbf{x})}(Archive)$  \Comment{Select $Point_j$ with the largest $f(\textbf{x})$ value in $Archive$}
\State  $PeakSet_n = \emptyset \cup {Point_j}$ 	\label{line_peak1_alg_peak_detection}
\State  $Archive =  Archive  \backslash  Point_j$   \Comment{Remove the point that has been selected}
\While  {$Archive \neq \emptyset$}
    \State  $Point_j = \max\limits_{f(\textbf{x})}(Archive)$  
    \State  $S = \{k\mid$ there are no smaller points in the neighborhood of $Point_j$ with regard to $PeakSet_k, k=1, ..., n\}$ \label{line_lower_y_alg_peak_detection}
    \If	{$S \neq \emptyset$} \label{line_empty_S_alg_peak_detection}
        \State  $k^{\ast} = \arg\min {Dist}(Point_j, PeakSet_k), k \in S$ 
        \State  $PeakSet_{k^\ast} = PeakSet_{k^\ast} \cup Point_j$ 
    \Else
        \If {$n<N_{p}$} 
            \State  $n = n+1$   
            \State  $PeakSet_n = \emptyset \cup Point_j$	\Comment{Record the peak in $PeakSet_n$} \label{line_new_peak_alg_peak_detection}
        \Else
            \State  \textbf{break} \Comment{Break if $N_{p}$ peaks have been found}
        \EndIf	
    \EndIf		
    \State  $Archive =  Archive  \backslash  Point_j$   
\EndWhile	
\State    $\Psi = \{Peak_k \mid$ find $Peak_k$ in $PeakSet_k$ according to Eq.(\ref{eq_peak}), $k=1,...,n\}$	
\State\Return	$\Psi$
\end{algorithmic}
\end{algorithm}
}

\subsection{Robust solution-searching stage}
After the peaks are detected in the peak-detection stage, we will move to the robust solution-searching stage.
The main purpose of this stage is to locate the robust optimal solution quickly based on the information of the peaks found.

In this stage, we use a global optimizer to search for the robust optimal solution of the problem.
The details of conducting the robust solution-searching procedure are shown in Algorithm \ref{alg_solution_searching}.
In Algorithm \ref{alg_solution_searching}, we utilize a differential evolution (DE) variant as the global optimizer due to its strong capability to handle complex optimization problems.
The mutation operator in the DE optimizer is DE/rand-to-\textit{p}best/1 (Line 7) \cite{zhang2009adaptive}, where $\textbf{x}_{rp,G}$ is randomly selected from the peaks obtained from the peak-detection stage (Line 5).
The purpose is to guide the population to search around the peaks, which are considered as the promising regions where the points are more likely to evolve into robust optimal solutions.
After the mutation operation, the individual undergoes the binomial crossover which is a common crossover operation of DE (Line 8).
Then the offspring will take the place of its parent if it has a larger mean effective objective value (Lines 10-12).
When the termination criterion is met, we return the individual with the maximal $f^{\textrm{eff}}(\textbf{x})$ value as the robust optimal solution of the problem.

{\linespread{1}
\begin{algorithm}[!h]
\small
\caption{$\textit{Global\_Optimizer}(Q_{0},\bm{\delta},\Psi)$}
\label{alg_solution_searching}
\hspace*{0.02in} {\bf Input:} $Q_0$ -- the initialized population  \\
\hspace*{0.4in} $\bm{\delta}$ --  the perturbation imposed on the decision vector\\
\hspace*{0.39in} $\Psi$ -- the peaks detected \\
\hspace*{0.02in} {\bf Output:}  $\textbf{x}^*$ -- the robust optimal solution
\begin{algorithmic}[1]
\State $G = 0$
\Comment{Initialize the generation counter}
\While  {$G < G_{max2}$}
  \State $Q_{G+1}=Q_G$
  \For {$i = 1 $ to $N$}
    \State Randomly select an individual $\textbf{x}_{rp}$ from $\Psi$
    \State Randomly select three individuals $\textbf{x}_{r_1,G},\textbf{x}_{r_2,G},\textbf{x}_{r_3,G}$ from $Q_G$, $r_1\neq r_2\neq r_3\neq i$
    \State $\textbf{v}_{i,G}=\textbf{x}_{r_1,G}+F\cdot(\textbf{x}_{rp,G}-\textbf{x}_{r_1,G})+F\cdot(\textbf{x}_{r_2,G}-\textbf{x}_{r_3,G})$
    \State Generate $\textbf{u}_{i,G}$ from $\textbf{x}_{i,G}$ and $\textbf{v}_{i,G}$ by binomial crossover
    \State Calculate $f^{\textrm{eff}}(\textbf{u}_{i,G})$ according to Eq. (\ref{eqprofi})
    \If {$f^{\textrm{eff}}(\textbf{u}_{i,G})>f^{\textrm{eff}}(\textbf{x}_{i,G}$)}
    \State $\textbf{x}_{i,G+1}=\textbf{u}_{i,G}$
    \EndIf
  \EndFor
  \State $\textbf{x}^*_{G+1}=\max\limits_{f^{\textrm{eff}}(\textbf{x})}\{\textbf{x}_{i,G+1}\in Q_{G+1}|i=1,2,...,N\}$
  \State $G=G+1$
\EndWhile
\State\Return	$\textbf{x}^*_{G}$
\end{algorithmic}\label{alg5}
\end{algorithm}
}

\subsection{Computational complexity of DREA}
Here, we discuss the computational complexity of DREA.
There are two main stages in DREA.
In the first stage, the main operations are \textit{Global\_Optimizer}, \textit{Archive\_Update} and \textit{Peak\_Detection}.
The computational complexity of \textit{Global\_Optimizer} depends on the optimizer we use.
For example, $\mathcal{O}(G_{max1}\cdot N\cdot D)$ for the original DE, where $G_{max1}$ is maximum iterations for the first stage, $N$ is the population size, $D$ is the dimension size.
The complexity of \textit{Archive\_Update} comes from selecting the point with the largest fitness value from $G_{max1}\cdot N$ points in $Archive$.
Hence the computational complexity of \textit{Archive\_Update} is $\mathcal{O}(G_{max1}\cdot N)$.
In \textit{Peak\_Detection}, we need to sort the points in $Archive$, the complexity of which is $\mathcal{O}(N_a\textrm{log}N_a)$, where $N_a$ is the archive size.
Therefore, it can be found that the computational complexity of the first stage mainly depends on the global optimizer we use.

In the second stage, the main operation is \textit{Global\_Optimizer}, where the perturbation is considered and the mean effective objective functions are computed by sampling the related $m$ neighbouring points.
Hence the computational complexity of \textit{Global\_Optimizer} depends on the optimizer used.
For example, $\mathcal{O}(G_{max2}\cdot N\cdot D\cdot m)$ for the original DE, where $G_{max2}$ is maximum iterations for the second stage.
Therefore, the computational complexity of DREA mainly depends on the global optimizer used in the two stages.

\section{Experimental Studies}
\subsection{Test problems}
We use six test problems in the experimental studies.
These test problems have complex characteristics that make them very challenging to solve \cite{du2022searching,lim2006inverse,mirjalili2019improving}.
The descriptions are as follows:
\begin{itemize}
  \item [$\bullet$]
  $f_1$: $f_1$ is a multimodal problem with a global optimum and a number of local optima, which is created using an expansion in terms of Gaussian basis functions.
  \item [$\bullet$]
  $f_2, f_3$: $f_2$ and $f_3$ have distinct deceptive search spaces, each containing a number of deceptive local optima alongside a true optimum. The shape of each search space is designed by different functions for each test problem, all tailored to favor the deceptive optima.
  \item [$\bullet$]
  $f_4, f_5$: $f_4$ and $f_5$ feature a multimodal search space. Each test problem has many non-robust global and local optima, each defined by a distinct function.
  \item [$\bullet$]
  $f_6$: $f_6$ exhibits a very flat search space, where the function values of the points are identical. The shape of the search space is crafted by a specific function.
\end{itemize}
Considering in real-world applications, perturbations are most likely to affect specific variables in the optimization problem \cite{du2018robust,he2019evolutionary,du2022searching}. This implies that a relatively low number of variables are subject to perturbations.
Therefore, in this paper, each test problem is extended to three different low dimensions: 10-$D$, 15-$D$ and 20-$D$.
Hence we have a total of 18 cases in the simulation studies.
The detailed formulations of the test problems are provided in Table \ref{testfun}.
The robust optimal solutions of these 18 problems are listed in Table \ref{funsolution}.
The related robust fitness values of these problems can be calculated analytically or approximated by Monte-Carlo integration.

\begin{table*}[!htbp]
\renewcommand\arraystretch{1.25}
\footnotesize
\centering
\begin{threeparttable}
\caption{Test problems used in this paper.}\label{testfun}
\begin{tabular}{c|l|c|c}
\shline
\hline
\textbf{Test Problem} & \textbf{Objective Function}\tnote{1} & \textbf{Search Space} & \textbf{Perturbation} \\\shline
\hline
\multirow{5}*{$f_{1}$} &&\\
& $f(\textbf{x})=\sum_{j=1}^{m}\left(\beta_{j} \prod_{i=1}^{d} e^{\frac{-\left(x_{i}-\mu_{j i}\right)^{2}}{2 \sigma_{j}^{2}}}\right).$ & $0\leq x_i\leq10$ & $\delta_i\sim U(-0.1,0.1)$ \\
&& \\
\hline
\multirow{5}*{$f_{2}$} & $f(\textbf{x}) = -\left(H\left(x_{1}\right)+H\left(x_{2}\right)\right) \times G(\textbf{x})+1,$ & &  \\
& $\textrm{where} \hspace{0.15cm} H(x)=\frac{1}{2}-0.3 e^{-\left(\frac{x-0.4}{0.004}\right)^{2}}$ & $0\leq x_i\leq1$ & $\delta_i\sim U(-0.01,0.01)$\\
&$\hspace{1.9cm} -0.5e^{-\left(\frac{x-0.5}{0.05}\right)^{2}}-0.3e^{-\left(\frac{x-0.6}{0.004}\right)^{2}}+\sin (\pi x),$ &\\
& $\hspace{0.84cm} G(\textbf{x})=\left(\sum_{i=3}^{N} 50 x_{i}^{2}\right)+1.$ & \\
\hline
\multirow{5}*{$f_{3}$} & $f(\textbf{x}) = -\left(H\left(x_{1}\right)+H\left(x_{2}\right)\right) \times G(\textbf{x})+1,$ & &  \\
& $\textrm{where} \hspace{0.15cm} H(x)=\frac{1}{2}-0.5e^{-\left(\frac{x-0.5}{0.05}\right)^{2}}$ & $0\leq x_i\leq1$ & $\delta_i\sim U(-0.01,0.01)$\\
&$\hspace{1.9cm}-\Sigma_{i=1}^{11}\left(0.3e^{-\left(\frac{x-0.04i}{0.004}\right)^{2}}+0.3\right.\left.e^{-\left(\frac{x-(1-0.04i)}{0.004}\right)^{2}}+\sin (\pi x)\right),$ &\\
\hline
\multirow{5}*{$f_{4}$} & $f(\textbf{x}) = -\left(H\left(x_{1}\right)+H\left(x_{2}\right)\right) \times G(\textbf{x})+1.399,$ & &  \\
& $\textrm{where} \hspace{0.15cm} H(x)=\frac{3}{2}-0.5 e^{-\left(\frac{x-0.5}{0.04}\right)^{2}} $ & $0\leq x_i\leq1$ & $\delta_i\sim U(-0.01,0.01)$\\
& $ \hspace{1.9cm} -\Sigma_{i=0}^{16}\left(0.8 e^{-\left(\frac{x-0.0063 i}{0.004}\right)^{2}}+0.8\right.\left.e^{-\left(\frac{x-(1-0.0063 i}{0.004}\right)^{2}}\right),$ &\\
& $\hspace{0.84cm} G(\textbf{x})=\left(\sum_{i=3}^{N} 50 x_{i}^{2}\right).$ & \\
\hline
\multirow{5}*{$f_{5}$} & $f(\textbf{x}) = -\left(H\left(x_{1}\right)+H\left(x_{2}\right)\right) \times G(\textbf{x})+1.399,$ & &  \\
& $\textrm{where} \hspace{0.15cm} H(x)=\frac{3}{2}-0.8e^{-\left(\frac{x-0.5}{0.04}\right)^{2}}$ & $0\leq x_i\leq1$ & $\delta_i\sim U(-0.01,0.01)$\\
& $ \hspace{1.9cm} -\Sigma_{i=0}^{16}\left(0.5e^{-\left(\frac{x-0.0063 i}{0.004}\right)^{2}}+0.5\right.\left.e^{-\left(\frac{x-(1-0.0063 i)}{0.004}\right)^{2}}\right),$ &\\
& $\hspace{0.84cm} G(\textbf{x})=\left(\sum_{i=3}^{N} 50 x_{i}^{2}\right).$ & \\
\hline
\multirow{5}*{$f_{6}$} & $f(\textbf{x}) = -\left(H\left(x_{1}\right)+H\left(x_{2}\right)\right) \times G(\textbf{x})+2,$ & &  \\
& $\textrm{where} \hspace{0.15cm} H(x)=\frac{1}{2}-\left(0.2 e^{-\left(\frac{x-0.95}{0.03}\right)^{2}}+0.2e^{-\left(\frac{x-0.05}{0.01}\right)^{2}}\right),$ & $0\leq x_i\leq1$ & $\delta_i\sim U(-0.01,0.01)$\\
& $\hspace{0.84cm} G(\textbf{x})=\left(\sum_{i=3}^{N} 50 x_{i}^{2}\right)+1.$ & \\
\shline
\end{tabular}
\begin{tablenotes}
\item[1] All the objective functions are to be maximized.
\end{tablenotes}
\end{threeparttable}
\end{table*}

\begin{table*}[htbp]
\footnotesize
	\centering
\caption{The robust optimal solutions and the original optimal solutions of the 18 cases.}\label{funsolution}
	\begin{tabular}{c|c|cc}
\shline
		\hline
		\textbf{Problem} & \textbf{Dimension} & \textbf{Robust Optimal Solution} & \textbf{Original Optimal Solution} \\\shline
		\hline
		\multirow{3}[0]{*}{$f_{1}$} & 10-$D$  & $(6,4,1.3,5,5,3,4,8,4,2)$ & $(6,4,1.3,5,5,3,4,8,4,2)$ \\\cline{2-4}
		& 15-$D$   & \tabincell{c}{$(6,4,1.3,5,5,3,4,$\\$8,4,2,1,3,5,7,9)$} & \tabincell{c}{$(6,4,1.3,5,5,3,4,$\\$8,4,2,1,3,5,7,9)$} \\\cline{2-4}
		& 20-$D$   & \tabincell{c}{$(6,4,1.3,5,5,3,4,8,4,2,$\\$1,3,5,7,2,1,3,5,7,1)$} & \tabincell{c}{$(6,4,1.3,5,5,3,4,8,4,2,$\\$1,3,5,7,2,1,3,5,7,1)$} \\
		\hline
\multirow{5}[0]{*}{$f_{2}$} & 10-$D$  & $(1,1,0,0,0,0,0,0,0,0)$ & $(0,0,0,0,0,0,0,0,0,0)$ \\\cline{2-4}
		& 15-$D$   & \tabincell{c}{$(1,1,0,0,0,0,0,$\\$0,0,0,0,0,0,0,0)$} & \tabincell{c}{$(0,0,0,0,0,0,0,$\\$0,0,0,0,0,0,0,0)$} \\\cline{2-4}
		& 20-$D$   & \tabincell{c}{$(1,1,0,0,0,0,0,0,0,0,$\\$0,0,0,0,0,0,0,0,0,0)$} & \tabincell{c}{$(0,0,0,0,0,0,0,0,0,0,$\\$0,0,0,0,0,0,0,0,0,0)$} \\
		\hline
\multirow{15}[0]{*}{$f_{3}$} &  &  & a total of 4 global optimal solutions: \\
& & & $(0,0,0,0,0,0,0,0,0,0)$ \\
&10-$D$  & $(0.5,0.5,0,0,0,0,0,0,0,0)$ & $(1,0,0,0,0,0,0,0,0,0)$ \\
& & & $(0,1,0,0,0,0,0,0,0,0)$ \\
& & & $(1,1,0,0,0,0,0,0,0,0)$ \\\cline{2-4}
		&    &  & a total of 4 global optimal solutions: \\
& & & $(0,0,0,0,0,0,0,0,0,0,0,0,0,0,0)$ \\
& 15-$D$ & $(0.5,0.5,0,0,0,0,0,0,0,0,0,0,0,0,0)$ & $(0,1,0,0,0,0,0,0,0,0,0,0,0,0,0)$ \\
& & & $(1,0,0,0,0,0,0,0,0,0,0,0,0,0,0)$ \\
& & & $(1,1,0,0,0,0,0,0,0,0,0,0,0,0,0)$ \\\cline{2-4}
		&    &  & a total of 4 global optimal solutions: \\
& & & $(0,0,0,0,0,0,0,0,0,0,0,0,0,0,0,0,0,0,0,0)$ \\
& 20-$D$ & $(0.5,0.5,0,0,0,0,0,0,0,0,0,0,0,0,0,0,0,0,0,0)$ & $(0,1,0,0,0,0,0,0,0,0,0,0,0,0,0,0,0,0,0,0)$ \\
& & & $(1,0,0,0,0,0,0,0,0,0,0,0,0,0,0,0,0,0,0,0)$ \\
& & & $(1,1,0,0,0,0,0,0,0,0,0,0,0,0,0,0,0,0,0,0)$ \\\cline{2-4}
		\hline
\multirow{15}[0]{*}{$f_{4}$} &  &  & a total of 4 global optimal solutions: \\
& & & $(0,0,0,0,0,0,0,0,0,0)$ \\
&10-$D$  & $(0.5,0.5,0,0,0,0,0,0,0,0)$ & $(1,0,0,0,0,0,0,0,0,0)$ \\
& & & $(0,1,0,0,0,0,0,0,0,0)$ \\
& & & $(1,1,0,0,0,0,0,0,0,0)$ \\\cline{2-4}
		&    &  & a total of 4 global optimal solutions: \\
& & & $(0,0,0,0,0,0,0,0,0,0,0,0,0,0,0)$ \\
& 15-$D$ & $(0.5,0.5,0,0,0,0,0,0,0,0,0,0,0,0,0)$ & $(0,1,0,0,0,0,0,0,0,0,0,0,0,0,0)$ \\
& & & $(1,0,0,0,0,0,0,0,0,0,0,0,0,0,0)$ \\
& & & $(1,1,0,0,0,0,0,0,0,0,0,0,0,0,0)$ \\\cline{2-4}
		&    &  & a total of 4 global optimal solutions: \\
& & & $(0,0,0,0,0,0,0,0,0,0,0,0,0,0,0,0,0,0,0,0)$ \\
& 20-$D$ & $(0.5,0.5,0,0,0,0,0,0,0,0,0,0,0,0,0,0,0,0,0,0)$ & $(0,1,0,0,0,0,0,0,0,0,0,0,0,0,0,0,0,0,0,0)$ \\
& & & $(1,0,0,0,0,0,0,0,0,0,0,0,0,0,0,0,0,0,0,0)$ \\
& & & $(1,1,0,0,0,0,0,0,0,0,0,0,0,0,0,0,0,0,0,0)$ \\\cline{2-4}
		\hline
\multirow{3}[0]{*}{$f_{5}$} & 10-$D$  & $(0.5,0.5,0,0,0,0,0,0,0,0)$ & a total of 1156 global optimal solutions \\\cline{2-4}
		& 15-$D$   & $(0.5,0.5,0,0,0,0,0,0,0,0,0,0,0,0,0)$ & a total of 1156 global optimal solutions \\\cline{2-4}
		& 20-$D$   & $(0.5,0.5,0,0,0,0,0,0,0,0,0,0,0,0,0,0,0,0,0,0)$ & a total of 1156 global optimal solutions \\
		\hline
\multirow{15}[0]{*}{$f_{6}$} &  &  & a total of 4 global optimal solutions: \\
& & & $(0.05,0.05,0,0,0,0,0,0,0,0)$ \\
&10-$D$  & $(0.95,0.95,0,0,0,0,0,0,0,0)$ & $(0.05,0.95,0,0,0,0,0,0,0,0)$ \\
& & & $(0.95,0.05,0,0,0,0,0,0,0,0)$ \\
& & & $(0.95,0.95,0,0,0,0,0,0,0,0)$ \\\cline{2-4}
		&    &  & a total of 4 global optimal solutions: \\
& & & $(0.05,0.05,0,0,0,0,0,0,0,0,0,0,0,0,0)$ \\
& 15-$D$ & $(0.95,0.95,0,0,0,0,0,0,0,0,0,0,0,0,0)$ & $(0.05,0.95,0,0,0,0,0,0,0,0,0,0,0,0,0)$ \\
& & & $(0.95,0.05,0,0,0,0,0,0,0,0,0,0,0,0,0)$ \\
& & & $(0.95,0.95,0,0,0,0,0,0,0,0,0,0,0,0,0)$ \\\cline{2-4}
		&    &  & a total of 4 global optimal solutions: \\
& & & $(0.05,0.05,0,0,0,0,0,0,0,0,0,0,0,0,0,0,0,0,0,0)$ \\
& 20-$D$ & $(0.95,0.95,0,0,0,0,0,0,0,0,0,0,0,0,0,0,0,0,0,0)$ & $(0.05,0.95,0,0,0,0,0,0,0,0,0,0,0,0,0,0,0,0,0,0)$ \\
& & & $(0.95,0.05,0,0,0,0,0,0,0,0,0,0,0,0,0,0,0,0,0,0)$ \\
& & & $(0.95,0.95,0,0,0,0,0,0,0,0,0,0,0,0,0,0,0,0,0,0)$ \\\cline{2-4}
		\hline
	\end{tabular}%
\end{table*}%

\subsection{Experimental settings}\label{secivb}
In DREA, two global optimizers are utilized across two distinct stages.
In our experiments, we implement the neighborhood-based crowding DE (NCDE) \cite{qu2012differential} during the peak-detection stage, aiming to efficiently locate both global and local peaks.
Subsequently, in the robust solution-searching stage, we employ DE/rand-to-\textit{p}best/1/bin to swiftly identify the robust optimal solution based on the information obtained from the peaks identified in the initial stage.
The scale factor $F$ and crossover rate $CR$ for both NCDE and DE/rand-to-\textit{p}best/1/bin are set to $F=0.5$ and $CR=0.9$.
The population size for both stages is fixed at $NP=100$.
The number of the peaks to detect is set as $N_p=3$, while the archive size is determined as $N_a=10000$, a size deemed ample for this research's purposes.
Moreover, the maximum fitness evaluations, $G_{max1}$ and $G_{max1}$, are set to different values based on the dimensions of the test problems, as detailed in Table \ref{maxfes}.
\begin{table}[htbp]
\footnotesize
\centering
\caption{The maximum fitness evaluations in each stage for the test problems with different dimensions.}\label{maxfes}
\begin{tabular}{c|c|c}
\shline
 & \tabincell{c}{\textbf{Peak-Detection} \\ \textbf{Stage}} & \tabincell{c}{\textbf{Robust Solution-Searching} \\ \textbf{Stage}} \\\shline
10-$D$ & $1\times10^4$ & $3\times10^5$  \\\hline
15-$D$ & $2\times10^4$ & $6\times10^5$  \\\hline
20-$D$ & $3\times10^4$ & $9\times10^5$  \\\shline
\end{tabular}
\end{table}

The algorithms used in comparison are JADE \cite{zhang2009jade}, CDE \cite{thomsen2004multimodal}, NCDE \cite{qu2012differential}, PRPSO \cite{mirjalili2016obstacles}, and RPSO\_IC \cite{mirjalili2019improving}.
JADE is a widely-used global optimizer, which belongs to a variant of differential evolution.
CDE and NCDE are two popular multimodal optimizers known for achieving superior performance in problems with multiple local optima.
PRPSO and RPSO\_IC are two recently proposed robust evolutionary algorithms.
Among them, PRPSO utilizes a penalty function to penalize solutions in proportion to their sensitivity to perturbations, while RPSO counts the number of previously sampled points in the neighbourhood of a solution and checks the reliability of a solution's robustness measure.
The parameter settings for these algorithms remain consistent with their original work.
The scale factor $F$ and the crossover rate $CR$ in CDE and NCDE are set as 0.5 and 0.9, respectively.
For JADE, $F$ and $CR$ are generated in an adaptive manner, which follows a Cauchy distribution and a normal distribution, respectively.
The location parameter $\mu_F$ and the mean $\mu_{CR}$ are initialized to be 0.5, as in \cite{zhang2009jade}.
The inertial weight $\omega$ of the particle swarm optimizer (PSO) in PRPSO and RPSO\_IC is linearly decreased from 0.9 to 0.4 over a specified range.
The acceleration constants $c_1$ and $c_2$ are both set to 2.
In RPSO\_IC, the number of sampled points $\zeta$ is set to $2\cdot D$, where $D$ is the dimensionality of the problem.
The population size is set as 100 for these algorithms.
The maximum fitness evaluations for these algorithms are set to be the same as those for DREA.

For all the algorithms, the number of the neighboring points used to compute the mean effective objective function is set as $H=100$.
All the algorithms are run 30 times in the experiments.
The algorithms are implemented with MATLAB R2021a and executed on Windows 10 with Intel(R) Xeon(R) W-10855M CPU @ 2.80 GHz 2.81 GHz and 32.0 GB RAM.

\subsection{Performance comparison on test problems}
We first compare the performance of DREA with five popular DE variants: CDE, NCDE, JADE, PBRO, and RPSO\_IC.
The results are listed in Table \ref{exp1}.
The results with a significant advantage (Wilcoxon's rank-sum test at a 0.05 significance level) are shown in bold.

From Table \ref{exp1}, it can be observed that the proposed DREA achieves the best performance among the six algorithms on all 18 test problems.
The promising performance of DREA is attributed to the different search behavior of the two stages.
The first stage of DREA aids in detecting the peaks of the original optimization problem's fitness landscape.
Subsequently, the second stage of DREA facilitates the quick localization of the robust optimal solution based on the peaks identified in the first stage.

\begin{figure*}
\centering
\subfigure[$f_1$ (10-$D$)]{
\begin{minipage}[b]{0.3\textwidth}
\includegraphics[width=1\textwidth]{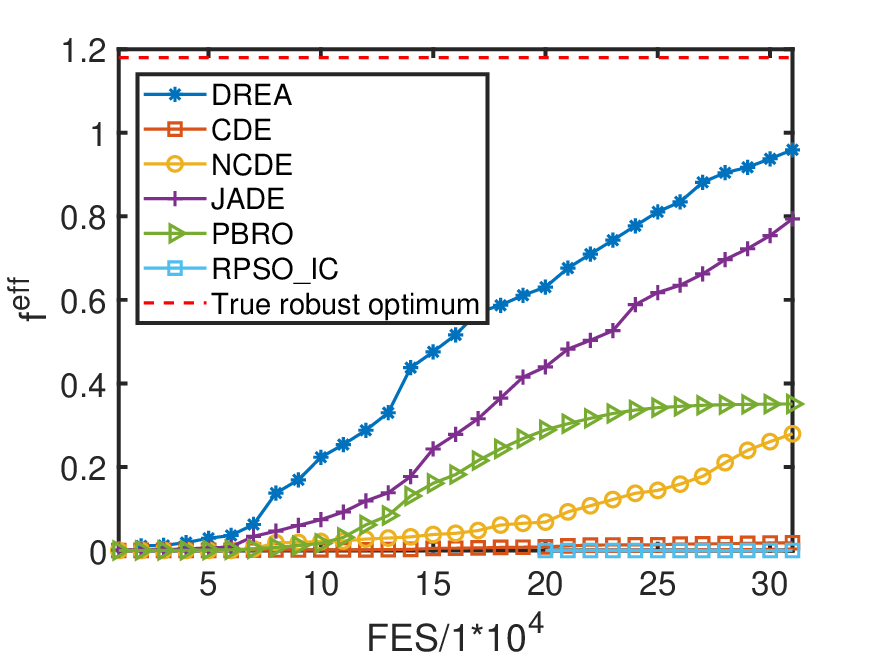}
\end{minipage}\label{2a}
}
\subfigure[$f_2$ (10-$D$)]{
\begin{minipage}[b]{0.3\textwidth}
\includegraphics[width=1\textwidth]{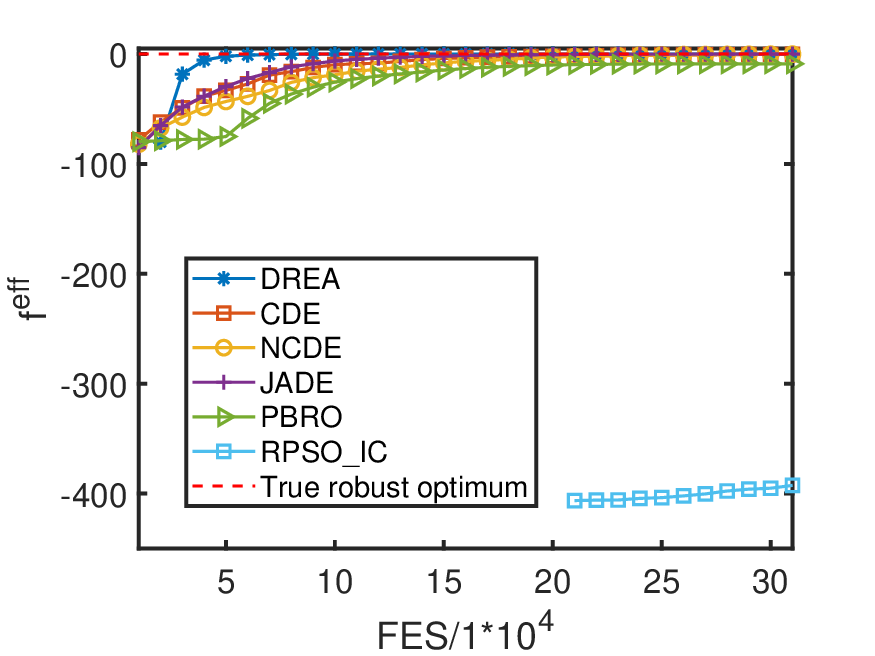}
\end{minipage}\label{2b}
}
\subfigure[$f_3$ (15-$D$)]{
\begin{minipage}[b]{0.3\textwidth}
\includegraphics[width=1\textwidth]{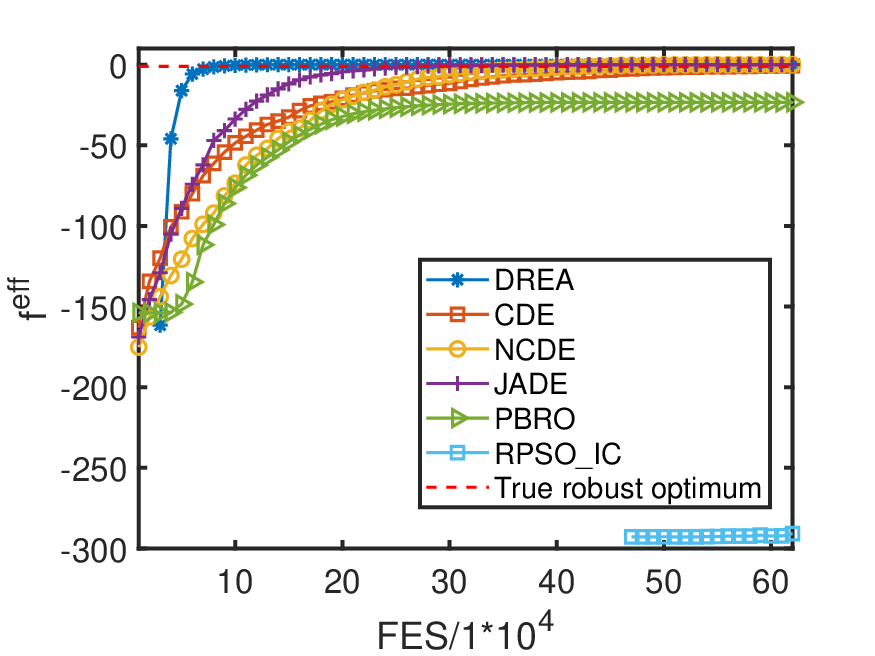}
\end{minipage}\label{2c}
}
\subfigure[$f_4$ (15-$D$)]{
\begin{minipage}[b]{0.3\textwidth}
\includegraphics[width=1\textwidth]{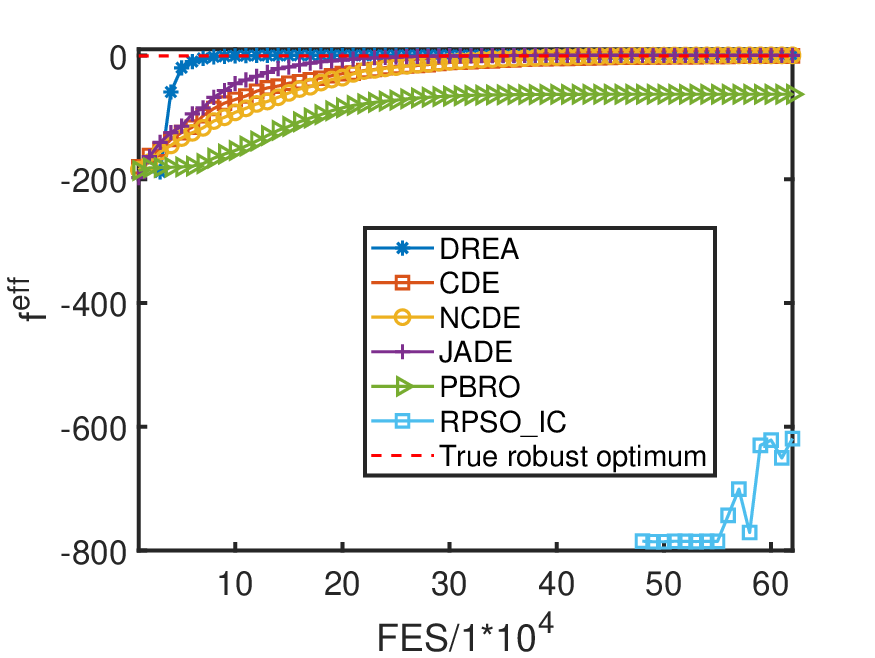}
\end{minipage}\label{2d}
}
\subfigure[$f_5$ (20-$D$)]{
\begin{minipage}[b]{0.3\textwidth}
\includegraphics[width=1\textwidth]{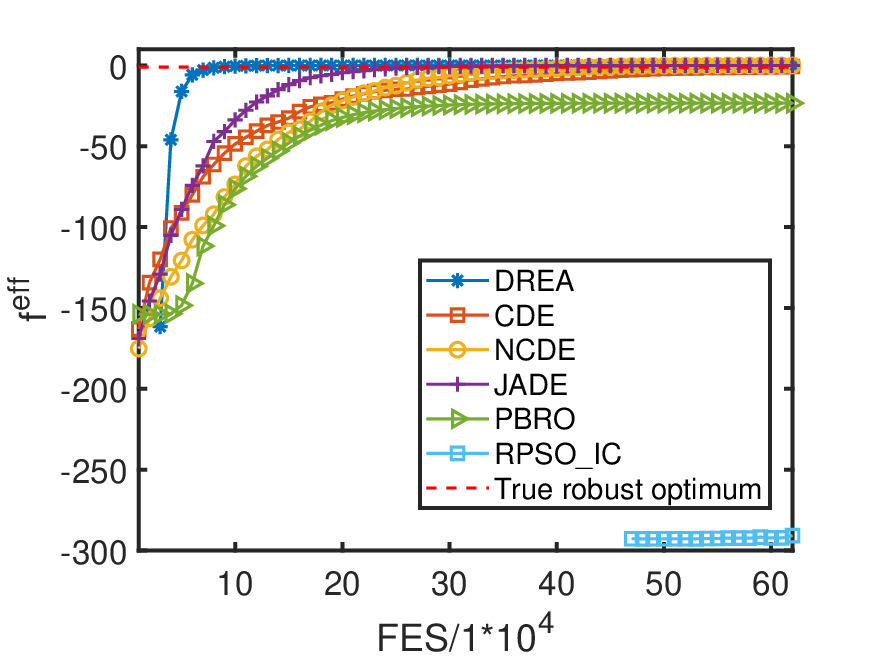}
\end{minipage}\label{2e}
}
\subfigure[$f_6$ (20-$D$)]{
\begin{minipage}[b]{0.3\textwidth}
\includegraphics[width=1\textwidth]{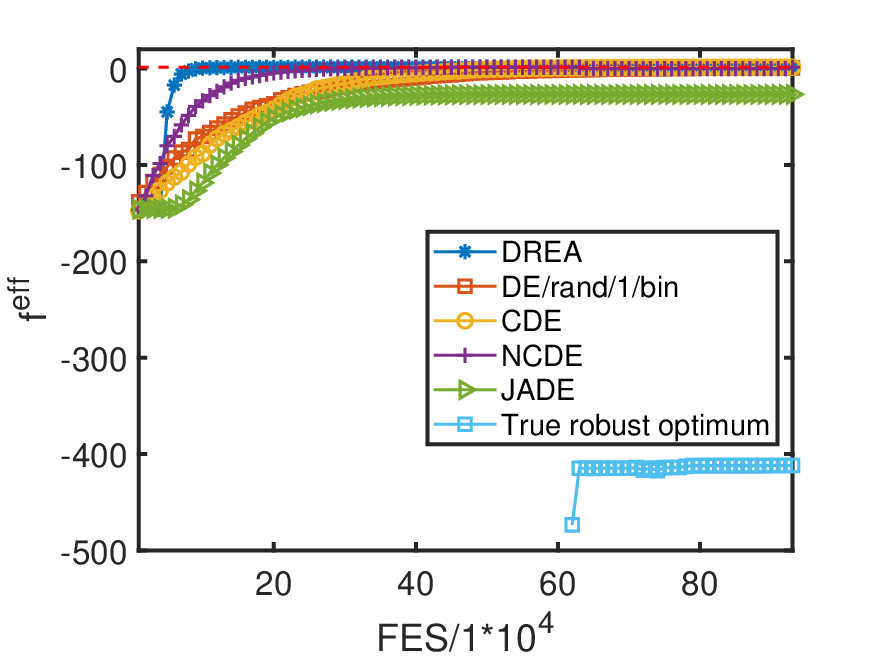}
\end{minipage}\label{2f}
}
\caption{The convergence curves derived from DREA, CDE, NCDE, JADE, PRPSO and RPSO\_IC on six test problems.} \label{fig2}
\end{figure*}

In Fig. \ref{fig2}, we present the convergence curves derived from DREA, CDE, NCDE, JADE, PBRO, and RPSO\_IC on the six selected test problems of varing dimensions.
It can be observed that using a small number of fitness evaluations in the first stage (i.e., $1\times10^4$ for 10-$D$ problems, $2\times10^4$ for 15-$D$ problems, $3\times10^4$ for 20-$D$ problems), DREA converges very rapidly for the test problems.
In addition to $f_1$ (10-$D$), DREA can converge to the true robust optima with a very limited number of fitness evaluations when compared to the other five algorithms.
However, it is also noted that RPSO\_IC failed to obtain feasible robust solutions at the early stage of the optimization process.
This is because RPSO\_IC imposes a constraint on computing the number of sampled points within the neighbourhood of a solution, which is difficult to satisfy during the early search.

Additionally, we recorded the running time of DREA and the other five algorithms on the 18 test problems in Table \ref{time}.
It is evident that the running time of DREA is much longer than that of the other five algorithms, primarily due to the computational overheads in the peak-detection stage.
However, the additional running time required by DREA is clearly justified by the higher quality of the solutions it produces.

\begin{table*}[htbp]
\footnotesize
	\centering
\caption{The experimental results of performance comparison on test problems.}\label{exp1}
	\begin{tabular}{cc|cc|cc|cc}
		\hline
		\multirow{2}[0]{*}{Dimension} & \multirow{2}[0]{*}{Problem} & \multicolumn{2}{c|}{DREA} &  \multicolumn{2}{c|}{CDE} & \multicolumn{2}{c}{NCDE} \\
		\cline{3-8}
		&       & Mean  & Std   & Mean  & Std   & Mean  & Std   \\
		\hline
		\multirow{6}[0]{*}{10-$D$} & $f_{1}$  & \textbf{9.59E-01} & \textbf{5.76E-02} &  1.86E-02$-$ & 1.48E-02 & 2.79E-01$-$ & 1.26E-01  \\
		& $f_{2}$     & \textbf{-1.33E-02} & \textbf{1.24E-05} & -3.09E-01$-$ & 3.87E-01 & -1.59E-01$-$ & 2.14E-01  \\
		& $f_{3}$     & \textbf{-1.40E-02} & \textbf{1.10E-03} & -2.92E-01$-$ & 3.55E-01 & -2.35E-01$-$ & 3.22E-01  \\
		& $f_{4}$     & \textbf{1.86E-01} & \textbf{6.76E-04} & -5.24E-01$-$ & 5.88E-01 & -3.18E-01$-$ & 4.60E-01  \\
		& $f_{5}$     & \textbf{-5.96E-02} & \textbf{5.50E-03} & -8.75E-01$-$ & 3.52E-01 & -9.33E-01$-$ & 6.62E-01  \\
		& $f_{6}$     & \textbf{1.37E+00} & \textbf{4.90E-03} & 9.37E-01$-$ & 1.91E-01 & 9.85E-01$-$ & 1.31E-01  \\
		\hline
		\multirow{6}[0]{*}{15-$D$} & $f_{1}$    & \textbf{1.02E+00} & \textbf{5.69E-02} & 1.62E-03$-$ & 1.70E-03 & 2.74E-01$-$ & 7.15E-02  \\
		& $f_{2}$     & \textbf{-2.16E-02} & \textbf{2.02E-07} & -8.61E-01$-$ & 8.61E-01 & -2.73E-02$-$ & 1.59E-02  \\
  		& $f_{3}$     & \textbf{-2.19E-02} & \textbf{1.40E-03} & -6.76E-01$-$ & 4.94E-01 & -2.86E-02$-$ & 1.39E-02  \\
		& $f_{4}$     & \textbf{1.77E-01} & \textbf{1.58E-04} & -8.29E-01$-$ & 7.42E-01 & 1.51E-01$-$ & 4.65E-02  \\
		& $f_{5}$     & \textbf{-6.50E-02} & \textbf{2.61E-04} & -1.08E+00$-$ & 5.31E-01 & -4.18E-01$-$ & 2.11E-01  \\
		& $f_{6}$     & \textbf{1.37E+00} & \textbf{3.25E-05} & 8.21E-01$-$ & 2.49E-01 & 1.23E+00$-$ & 8.49E-02  \\
		\hline
		\multirow{6}[0]{*}{20-$D$} & $f_{1}$    & \textbf{1.06E+00} & \textbf{7.89E-02} & 3.89E-05$-$ & 9.45E-05 & 2.38E-01$-$ & 9.00E-02  \\
		& $f_{2}$     & \textbf{-2.99E-02} & \textbf{1.58E-05} & -1.49E+00$-$ & 8.34E-01 & -3.00E-02$-$ & 4.56E-05  \\
		& $f_{3}$     & \textbf{-3.00E-02} & \textbf{3.34E-05} & -1.94E+00$-$ & 1.23E+00 & -3.00E-02$-$ & 6.43E-05  \\
		& $f_{4}$     & \textbf{1.67E-01} & \textbf{8.15E-05} & -1.89E+00$-$ & 1.57E+00 & 1.66E-01$-$ & 1.46E-03  \\
		& $f_{5}$     & \textbf{-7.67E-02} & \textbf{3.48E-05} & -1.44E+00$-$ & 5.75E-01 & -2.28E-01$-$ & 1.10E-01  \\
		& $f_{6}$     & \textbf{1.37E+00} & \textbf{8.93E-08} & 7.40E-01$-$ & 2.83E-01 & 1.33E+00$-$ & 2.62E-02  \\
		\hline
		\multicolumn{2}{c|}{$+$/$\approx$/$-$} & \multicolumn{2}{c|}{-} & \multicolumn{2}{c|}{0/0/18} &\multicolumn{2}{c}{0/0/18} \\
		\hline
\multirow{2}[0]{*}{Dimension} & \multirow{2}[0]{*}{Problem} & \multicolumn{2}{c|}{JADE} &  \multicolumn{2}{c|}{PRPSO} & \multicolumn{2}{c}{RPSO\_IC} \\
		\cline{3-8}
		&       & Mean  & Std   & Mean  & Std   & Mean  & Std  \\
		\hline
		\multirow{6}[0]{*}{10-$D$} & $f_{1}$  & 7.94E-01 & 1.05E-01 &  3.51E-01$-$ & 3.13E-01 & 1.86E-16$-$ & 8.54E-16  \\
		& $f_{2}$     & -6.60E-02 & 3.00E-02 & -8.92E+00$-$ & 1.05E+01 & -3.90E+02$-$ & 1.97E+02  \\
		& $f_{3}$     & -6.34E-02 & 2.45E-02 & -1.13E+01$-$ & 1.45E+01 & -3.12E+02$-$ & 1.04E+02  \\
		& $f_{4}$     & 1.62E-01 & 1.38E-02 & -3.29E+01$-$ & 3.04E+01 & -3.79E+02$-$ & 1.44E+02  \\
		& $f_{5}$     & -2.93E-01 & 7.73E-02 & -1.26E+01$-$ & 1.84E+01 & -5.10E+02$-$ & 1.70E+02  \\
		& $f_{6}$     & 1.30E+00 & 4.82E-02 & -1.56E+00$-$ & 3.37E+00 & -1.99E+02$-$ & 5.07E+01  \\
		\hline
		\multirow{6}[0]{*}{15-$D$} & $f_{1}$    & 6.43E-01 & 3.58E-01 & 5.83E-02$-$ & 1.22E-01 & 1.03E-24$-$ & 5.62E-24  \\
		& $f_{2}$     & -2.46E-02 & 3.36E-03 & -3.69E+01$-$ & 3.37E+01 & -5.82E+02$-$ & 1.82E+02  \\
  		& $f_{3}$     & -3.16E-02 & 6.92E-03 & -2.34E+01$-$ & 3.00E+01 & -6.00E+02$-$ & 2.25E+02  \\
		& $f_{4}$     & 1.73E-01 & 3.47E-03 & -6.27E+01$-$ & 6.15E+01 & -7.28E+02$-$ & 3.38E+02  \\
		& $f_{5}$     & -1.84E-01 & 8.89E-02 & -6.40E+01$-$ & 5.38E+01 & -8.57E+02$-$ & 2.25E+02  \\
		& $f_{6}$     & 1.35E+00 & 2.35E-02 & -8.46E+00$-$ & 1.41E+01 & -3.46E+02$-$ & 7.29E+01  \\
		\hline
		\multirow{6}[0]{*}{20-$D$} & $f_{1}$    & 4.90E-01 & 3.52E-01 & 5.27E-03$-$ & 9.63E-03 & 4.41E-38$-$ & 2.41E-37  \\
		& $f_{2}$     & -6.64E-02 & 2.28E-02 & -7.24E+01$-$ & 4.23E+01 & -7.87E+02$-$ & 3.49E+02  \\
		& $f_{3}$     & -8.32E-02 & 3.98E-02 & -6.28E+01$-$ & 3.78E+01 & -8.25E+02$-$ & 3.21E+02  \\
		& $f_{4}$     & 9.93E-02 & 5.67E-02 & -1.16E+02$-$ & 7.20E+01 & -9.86E+02$-$ & 3.41E+02  \\
		& $f_{5}$     & -4.31E-01 & 9.91E-02 & -9.84E+01$-$ & 6.16E+01 & -1.21E+03$-$ & 2.96E+02  \\
		& $f_{6}$     & 1.32E+00 & 2.82E-02 & -2.69E+01$-$ & 2.67E+01 & -4.84E+02$-$ & 8.01E+01  \\
		\hline
		\multicolumn{2}{c|}{$+$/$\approx$/$-$} & \multicolumn{2}{c|}{-} & \multicolumn{2}{c|}{0/0/18} &\multicolumn{2}{c}{0/0/18} \\
		\hline
	\end{tabular}%
\end{table*}%

\begin{table*}[htbp]
\footnotesize
	\centering
\caption{The running time (in seconds) of each algorithm on test problems.}\label{time}
	\begin{tabular}{cc|c|c|c|c|c|c}
		\hline
		Dimension & Problem & DREA & CDE & NCDE & JADE & PRPSO & RPSO\_IC \\
		\cline{3-8}
		\hline
		\multirow{6}[0]{*}{10-$D$} & $f_{1}$  & 1.9027 & 0.8168 & 0.8414 & 0.4587 & 0.2429 & 0.1306 \\
        & $f_{2}$  & 0.9983 & 0.5541 & 0.5815 & 0.2001 & 0.1105 & 0.1215 \\
        & $f_{3}$  & 1.1965 & 0.7343 & 0.7797 & 0.3785 & 0.1290 & 0.1204 \\
        & $f_{4}$  & 1.4498 & 1.0224 & 1.0724 & 0.6405 & 0.1772 & 0.1262 \\
        & $f_{5}$  & 1.4171 & 1.0117 & 1.0469 & 0.6099 & 0.1747 & 0.1288 \\
        & $f_{6}$  & 1.0004 & 0.5360 & 0.5626 & 0.1879 & 0.1000 & 0.1254 \\
		\hline
		\multirow{6}[0]{*}{15-$D$} & $f_{1}$  & 4.6680 & 3.2310 & 3.2823 & 1.9350 & 0.8527  & 0.5221 \\
        & $f_{2}$  & 1.8156 & 1.1288 & 1.1717 & 0.3899 & 0.2444 & 0.5127 \\
        & $f_{3}$  & 2.4385 & 1.5105 & 1.5672 & 0.7677 & 0.2905 & 0.5336 \\
        & $f_{4}$  & 2.7852 & 2.0753 & 2.1884 & 1.3518 & 0.4028 & 0.5196 \\
        & $f_{5}$  & 2.6708 & 2.0644 & 2.1073 & 1.2455 & 0.3882 & 0.5196 \\
        & $f_{6}$  & 1.8170 & 1.0930 & 1.1615 & 0.3756 & 0.2421 & 0.5414 \\
		\hline
		\multirow{6}[0]{*}{20-$D$} & $f_{1}$  & 9.2040 & 7.9637 & 8.1464 & 3.3237 & 1.8950 & 1.2982 \\
        & $f_{2}$  & 2.6691 & 1.6978 & 1.8482 & 0.3920 & 0.4464 & 1.2771 \\
        & $f_{3}$  & 4.1518 & 2.2727 & 2.4908 & 0.7515 & 0.5254 & 1.3429 \\
        & $f_{4}$  & 3.8427 & 3.1932 & 3.3943 & 1.3137 & 0.6878 & 1.4184 \\
        & $f_{5}$  & 3.7929 & 3.1043 & 3.2508 & 1.2272 & 0.6642 & 1.3121 \\
        & $f_{6}$  & 2.7468 & 1.6555 & 1.7993 & 0.3776 & 0.4531 & 1.3084 \\
		\hline
	\end{tabular}%
\end{table*}%

\subsection{Search behavior of DREA}
\begin{figure}[!ht]
\centering
\subfigure[$f_1$ (10-$D$)]{\includegraphics[width=8cm]{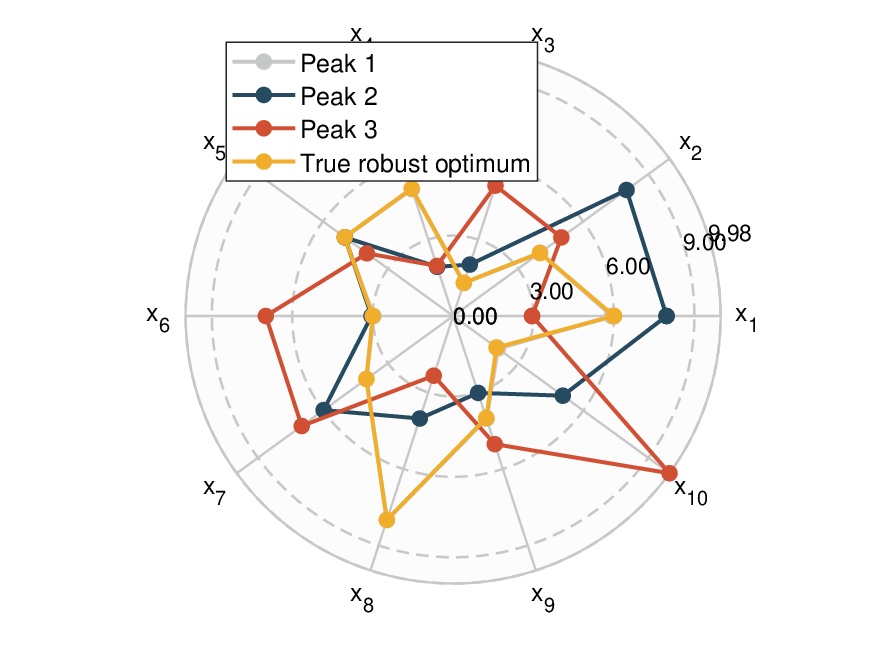}}\label{f1}
\subfigure[$f_4$ (20-$D$)]{\includegraphics[width=8cm]{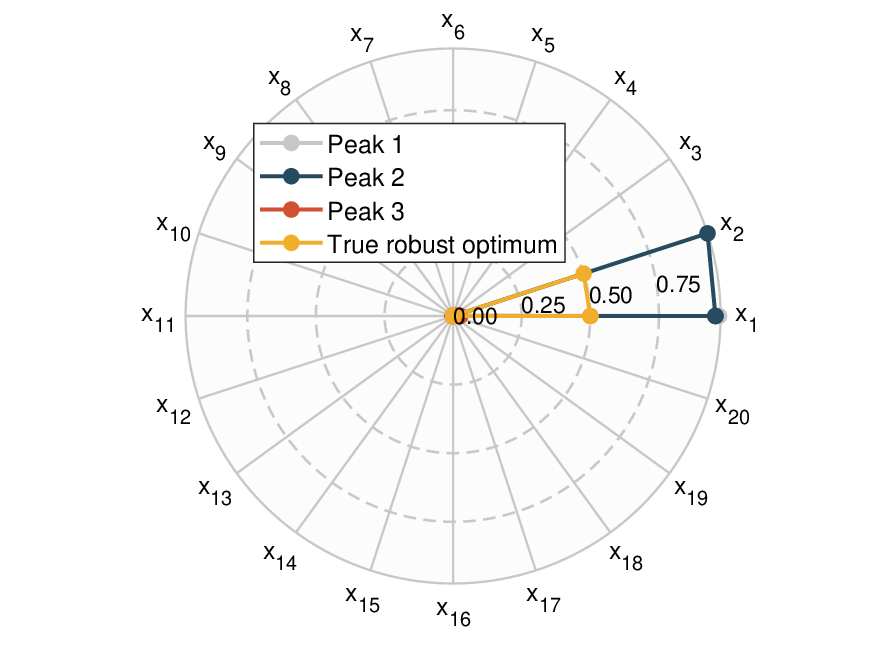}}\label{f4}
\caption{The peaks found by DREA in the first stage of $f_1$ (10-$D$) and $f_4$ (20-$D$).
}
\label{fig_behavior}
\end{figure}

In this subsection, we investigate the search behavior of the proposed DREA.
We select two test problems $f_1$ (10-$D$) and $f_4$ (20-$D$) as representative cases.
Fig. \ref{fig_behavior} displays the location of the peaks found by DREA in the first stage, represented in a radar chart.
The peaks are indicated by grey, dark blue and red dot-dash lines.
The true robust optimum of the problem is represented by the yellow dot-dash line.
The peaks obtained in these two test problems are also listed in Table \ref{peaks}.

In Fig. \ref{f1}, for $f_1$, the robust optimal solution coincides with the original optimal solution. The peak-detection stage successfully covers an area very close to the optimal region of the problem (see Peak1 of $f_1$ in Table \ref{peaks}).
In Fig. \ref{f4}, for $f_4$, the robust optimal solution differs from the original optimal solution.
Although the peak-detection stage only identifies the original optimal solutions (see Peaks 1-3 of $f_4$ in Table \ref{peaks}), these promising regions efficiently contribute to finding the robust optimal solution.
This operation has the potential to reduce the time needed to find the robust optimal solution, as reflected in Fig. \ref{fig2}.

\begin{table*}[htbp]
\footnotesize
\centering
\caption{The peaks found in the peak-detection stage.}\label{peaks}
\begin{tabular}{c|c}
\shline
 \textbf{Test Problem} & \textbf{Peaks Found at the Peak-Detection Stage} \\\shline
\multirow{3}[0]{*}{$f_1$ (10-$D$)} & Peak1: (5.92,4.01,1.29,4.97,5.00,2.96,4.00,8.00,4.03,2.07)  \\
 & Peak2: (7.96,8.00,2.02,1.93,4.99,3.05,5.97,4.02,3.02,5.06)  \\
 & Peak3: (2.94,4.99,5.11,1.96,3.98,6.99,6.98,2.34,5.03,9.98)  \\\hline
\multirow{3}[0]{*}{$f_4$ (20-$D$)} & Peak1: (0.97,0.01,1.66E-04,0,0,0,0,0,0,1.79E-04,0,2.33E-04,1.67E-04,0,0,0,0,0,2.05E-04,0) \\
 & Peak2: (0.96,0.97,0,0,6.70E-04,0,4.49E-04,0,4.55E-04,6.26E-04,0,1.28E-04,0,2.11E-04,0,0,0,0,3.32E-04,0) \\
 & Peak3: (0.03,0.006,1.74E-04,4.21E-04,0,0,0,0,0,7.56E-04,1.01E-04,9.35E-04,0,3.73E-04,0,2.80E-04,0,0,0,2.33E-04) \\\shline
\end{tabular}
\begin{tablenotes}
\item[1] Any variant value of the peaks smaller than 1E-04 is recorded as 0.
\end{tablenotes}
\end{table*}

\subsection{Parameter sensitivity analysis}
As mentioned in Section \ref{seciiib}, a number of peaks of the fitness landscape of the original optimization problem are detected in the first stage.
Then these peaks are used to guide the search in the second stage.
The number of the peaks to detect is set as $N_p=3$ in the experiments.
In this subsection, we investigate the sensitivity of $N_p$ on the total 18 test problems using five different $N_p$ values: 1, 2, 3, 4 and 5.
The robust optimal solutions of the 18 test problems found by DREA using different $N_p$ values are provided in Table \ref{exp2}.

We calculate the average rankings of the five different $N_p$ values and show them in the last row of Table \ref{exp2}, where a lower ranking indicates better performance.
It can be found that for the problems that the robust and the global optimal solutions are the same (i.e., $f_1$,$f_2$ and $f_5$), a smaller number of peaks has a better effect on the performance of DREA.
This is because a smaller number of peaks means the global optimal solution will have a larger chance to be selected as $\textbf{x}_{rp}$ in the global optimizer of Algorithm \ref{alg5}, which makes DREA with smaller $N_p$ values more efficient for these problems.
While for the problems that the robust and the global optimal solutions are different (i.e., $f_3$,$f_4$ and $f_6$), a larger number of peaks has a better effect on the performance of DREA.
This is because a larger number of peaks implies that more peaks will guide the search of the global optimizer in DREA, which enhances the diversity of the search.
According to the results in \ref{exp2}, it can be found that DREA with $N_p=3$ has promising performance on both types of test problems.
Therefore, we use $N_p=3$ for DREA in the experiments in this paper.

\begin{table*}[htbp]
\scriptsize
	\centering
\caption{The experimental results of the parameter sensitivity.}\label{exp2}
	\begin{tabular}{cc|cc|cc|cc|cc|cc}
		\hline
		\multirow{2}[0]{*}{Dimension} & \multirow{2}[0]{*}{Problem} & \multicolumn{2}{c|}{$N_p=1$} & \multicolumn{2}{c|}{$N_p=2$} & \multicolumn{2}{c|}{$N_p=3$} & \multicolumn{2}{c|}{$N_p=4$} & \multicolumn{2}{c}{$N_p=5$} \\
		\cline{3-12}
		&       & \tabincell{c}{Mean\\(Ranking)}  & Std   & \tabincell{c}{Mean\\(Ranking)}  & Std   & \tabincell{c}{Mean\\(Ranking)}  & Std   & \tabincell{c}{Mean\\(Ranking)}  & Std   & \tabincell{c}{Mean\\(Ranking)}  & Std \\
		\hline
		\multirow{6}[0]{*}{10-$D$} & $f_{1}$  & 1.17E+00(1) & 2.26E-03 & 1.13E+00(2) & 1.33E-02 & 9.59E-01(3) & 5.76E-02 & 9.53E-01(4) & 7.28E-02 & 8.94E-01(5) & 1.13E-01 \\
		& $f_{2}$     & -1.33E-02(3) & 1.83E-05 & -1.33E-02(1) & 2.99E-06 & -1.33E-02(2) & 1.24E-05 & -1.33E-02(4) & 1.92E-05 & -1.33E-02(5) & 2.59E-05 \\
		& $f_{3}$     & -2.65E-02(5) & 5.14E-03 & -1.80E-02(4) & 4.24E-03 & -1.40E-02(2) & 1.10E-03 & -1.37E-02(1) & 5.30E-04 & -1.41E-02(3) & 1.27E-03 \\
		& $f_{4}$     & 1.87E-01(1) & 3.45E-04 & 1.86E-01(2) & 3.34E-04 & 1.86E-01(3) & 6.76E-04 & 1.86E-01(5) & 7.80E-04 & 1.86E-01(4) & 8.94E-04 \\
		& $f_{5}$     & -7.81E-02(5) & 6.71E-02 & -5.52E-02(1) & 2.74E-03 & -5.96E-02(2) & 5.50E-03 & -6.01E-02(3) & 4.35E-03 & -6.25E-02(4) & 5.95E-03 \\
		& $f_{6}$     & 1.35E+00(5) & 2.99E-02 & 1.37E+00(2) & 1.33E-02 & 1.37E+00(1) & 4.90E-03 & 1.37E+00(3) & 4.73E-03 & 1.37E+00(4) & 1.04E-02 \\
		\hline
		\multirow{6}[0]{*}{15-$D$} & $f_{1}$    & 1.16E+00(1) & 7.30E-03 & 1.15E+00(2) & 1.16E-02 & 1.02E+00(5) & 5.69E-02 & 1.03E+00(4) & 6.05E-02 & 1.04E+00(3) & 5.83E-02 \\
		& $f_{2}$     & -2.16E-02(5) & 3.48E-05 & -2.16E-02(4) & 5.52E-06 & -2.16E-02(3) & 2.02E-07 & -2.16E-02(1) & 3.15E-08 & -2.16E-02(2) & 7.35E-08 \\
  		& $f_{3}$     & -3.55E-02(5) & 2.91E-03 & -2.43E-02(4) & 3.32E-03 & -2.19E-02(3) & 1.40E-03 & -2.17E-02(2) & 4.39E-05 & -2.17E-02(1) & 4.03E-05 \\
		& $f_{4}$     & 1.77E-01(1) & 2.04E-04 & 1.77E-01(2) & 1.33E-04 & 1.77E-01(3) & 1.58E-04 & 1.77E-01(5) & 1.85E-04 & 1.77E-01(4) & 2.12E-04 \\
		& $f_{5}$     & -6.55E-02(5) & 1.14E-03 & -6.49E-02(1) & 1.73E-04 & -6.50E-02(2) & 2.61E-04 & -6.50E-02(3) & 2.06E-04 & -6.51E-02(4) & 2.87E-04 \\
		& $f_{6}$     & 1.35E+00(5) & 2.54E-02 & 1.37E+00(4) & 1.54E-02 & 1.37E+00(1) & 3.25E-05 & 1.37E+00(2) & 4.28E-05 & 1.37E+00(3) & 8.78E-05 \\
		\hline
		\multirow{6}[0]{*}{20-$D$} & $f_{1}$    & 1.14E+00(2) & 1.26E-02 & 1.15E+00(1) & 3.82E-03 & 1.06E+00(3) & 7.89E-02 & 9.86E-01(4) & 6.29E-02 & 9.59E-01(5) & 5.89E-02 \\
		& $f_{2}$     & -3.00E-02(5) & 4.12E-05 & -3.00E-02(4) & 2.92E-05 & -2.99E-02(3) & 1.58E-05 & -2.99E-02(1) & 1.16E-08 & -2.99E-02(2) & 1.65E-08 \\
		& $f_{3}$     & -4.43E-02(5) & 2.71E-03 & -3.33E-02(4) & 3.83E-03 & -3.00E-02(3) & 3.34E-05 & -3.00E-02(1) & 2.43E-05 & -3.00E-02(2) & 2.85E-05 \\
		& $f_{4}$     & 1.67E-01(2) & 8.67E-05 & 1.67E-01(1) & 1.11E-04 & 1.67E-01(3) & 8.15E-05 & 1.67E-01(4) & 1.26E-04 & 1.67E-01(5) & 1.36E-04 \\
		& $f_{5}$     & -7.68E-02(5) & 3.63E-04 & -7.67E-02(2) & 6.66E-05 & -7.67E-02(1) & 3.48E-05 & -7.67E-02(3) & 6.19E-05 & -7.67E-02(4) & 8.24E-05 \\
		& $f_{6}$     & 1.33E+00(5) & 3.04E-02 & 1.36E+00(4) & 1.13E-02 & 1.37E+00(1) & 8.93E-08 & 1.37E+00(2) & 7.14E-07 & 1.37E+00(3) & 2.97E-06 \\
		\hline
		\multicolumn{2}{c|}{Average Ranking} & \multicolumn{2}{c|}{3.67} & \multicolumn{2}{c|}{2.50} & \multicolumn{2}{c|}{2.44} &\multicolumn{2}{c|}{2.89} & \multicolumn{2}{c}{3.50}\\
		\hline
	\end{tabular}%
\end{table*}%

\subsection{Scalability Study}
In this subsection, we investigate the performance of the proposed DREA on higher-dimensional robust optimization problems.
The test problems $f_2$ to $f_6$ in Table \ref{testfun} are extended to 100 and 200 dimensions.
It should be mentioned that we do not use $f_1$ because the robust optimal solution of $f_1$ is unknown when it is extended to higher-dimensions.
The results are provided in Table \ref{exps1}.

According to the results, it can be found that DREA outperforms all the five counterpart algorithms in comparison when the test problems are extended to 100-$D$ and 200-$D$.
However, as the dimensionality of the problem increases (from 100-$D$ to 200-$D$), the differences between DREA and some other algorithms (e.g., NCDE and JADE) become less evident.
In addition, the algorithms (CDE, NCDE and JADE) which use DE as the optimizer outperform the ones that use PSO (PRPSO and RPSO\_IC) as the optimizer.
This is because the superior performance of DE when handling high-dimensional problems.

Additionally, it should be noted that RPSO\_IC is unable to achieve feasible robust solutions as the dimensionality increases to 200.
This is because RPSO\_IC sets a constraint for calculating the number of sampled points within the neighbourhood of a solution, which is difficult to satisfy when the problem becomes higher-dimensional.

\begin{table*}[htbp]
\footnotesize
	\centering
\caption{The experimental results of performance comparison on higher-dimensional test problems.}\label{exps1}
	\begin{tabular}{cc|cc|cc|cc}
		\hline
		\multirow{2}[0]{*}{Dimension} & \multirow{2}[0]{*}{Problem} & \multicolumn{2}{c|}{DREA} &  \multicolumn{2}{c|}{CDE} & \multicolumn{2}{c}{NCDE} \\
		\cline{3-8}
		&       & Mean  & Std   & Mean  & Std   & Mean  & Std   \\
		\hline
		\multirow{6}[0]{*}{100-$D$} & $f_{2}$  & \textbf{-1.63E-01} & \textbf{2.92E-08} &  -2.50E+02$-$ & 3.34E+01 & -1.63E-01$-$ & 2.16E-04  \\
		& $f_{3}$     & \textbf{-1.63E-01} & \textbf{4.52E-06} & -2.50E+02$-$ & 2.80E+01 & -1.63E-01$-$ & 1.02E-04  \\
		& $f_{4}$     & 6.88E-03 & 2.15E-03 & -5.68E+02$-$ & 3.88E+01 & \textbf{8.38E-03}$+$ & \textbf{3.29E-04}  \\
		& $f_{5}$     & \textbf{-2.68E-01} & \textbf{1.89E-08} & -2.47E+02$-$ & 2.15E+01 & -2.71E-01$-$ & 1.93E-03  \\
		& $f_{6}$     & 1.29E+00 & 1.01E-08 & -9.84E+01$-$ & 1.32E+01 & 1.28E+00$-$ & 6.14E-04  \\
		\hline
		\multirow{6}[0]{*}{200-$D$} & $f_{2}$    & \textbf{-3.30E-01} & \textbf{3.92E-05} & -1.05E+03$-$ & 5.45E+01 & -3.33E-01$-$ & 5.76E-03  \\
		& $f_{3}$     & -3.48E-01 & 7.12E-03 & -1.05E+03$-$ & 6.41E+01 & \textbf{-3.33E-01}$+$ & \textbf{3.21E-03}  \\
  		& $f_{4}$     & -2.32E-01 & 4.48E-02 & -1.66E+03$-$ & 9.31E+01 & \textbf{-1.95E-01}$+$ & \textbf{4.22E-03}  \\
		& $f_{5}$     & -5.07E-01 & 7.30E-07 & -1.24E+03$-$ & 6.22E+01 & -5.52E-01$-$ & 1.37E-02  \\
		& $f_{6}$     & 1.15E+00 & 3.63E-02 & -5.27E+02$-$ & 2.65E+01 & 1.17E+00$\approx$ & 4.93E-03  \\
		\hline
		\multicolumn{2}{c|}{$+$/$\approx$/$-$} & \multicolumn{2}{c|}{-} & \multicolumn{2}{c|}{0/0/10} &\multicolumn{2}{c}{3/1/6} \\
		\hline
\multirow{2}[0]{*}{Dimension} & \multirow{2}[0]{*}{Problem} & \multicolumn{2}{c|}{JADE} &  \multicolumn{2}{c|}{PRPSO} & \multicolumn{2}{c}{RPSO\_IC}\tnote{1} \\
		\cline{3-8}
		&       & Mean  & Std   & Mean  & Std   & Mean  & Std   \\
		\hline
		\multirow{6}[0]{*}{100-$D$} & $f_{2}$  & -1.66E-01$-$ & 1.15E-02 &  -6.09E+02$-$ & 2.52E+02 & - & -  \\
		& $f_{3}$     & -1.69E-01$-$ & 1.42E-02 & -6.57E+02$-$ & 2.73E+02 & - & -  \\
		& $f_{4}$     & 2.12E-03$\approx$ & 1.60E-02 & -5.60E+02$-$ & 2.16E+02 & - & -  \\
		& $f_{5}$     & -2.69E-01$-$ & 2.80E-03 & -9.55E+02$-$ & 4.10E+02 & - & -  \\
		& $f_{6}$     & \textbf{1.29E+00}$+$ & \textbf{1.06E-07} & -4.43E+02$-$ & 1.92E+02 & - & -  \\
		\hline
		\multirow{6}[0]{*}{200-$D$} & $f_{2}$    & -3.62E-01$-$ & 8.89E-02 & -1.51E+03$-$ & 8.07E+02 & - & -  \\
		& $f_{3}$     & -3.68E-01$-$ & 1.45E-01 & -1.45E+03$-$ & 7.21E+02 & - & -  \\
  		& $f_{4}$     & -2.55E-01$\approx$ & 1.29E-01 & -1.43E+03$-$ & 4.63E+02 & - & -  \\
		& $f_{5}$     & \textbf{-5.07E-01}$+$ & \textbf{3.13E-06} & -2.45E+03$-$ & 8.99E+02 & - & -  \\
		& $f_{6}$     & \textbf{1.18E+00}$+$ & \textbf{3.77E-09} & -8.82E+02$-$ & 3.02E+02 & - & -  \\
		\hline
		\multicolumn{2}{c|}{$+$/$\approx$/$-$} & \multicolumn{2}{c|}{3/2/5} & \multicolumn{2}{c|}{0/0/10} &\multicolumn{2}{c}{0/0/10} \\
		\hline
	\end{tabular}%
\begin{tablenotes}
\item[1] ``-'' denotes that no feasible robust solutions are found.
\end{tablenotes}
\end{table*}%

\section{Conclusion}
In this paper, we propose a novel robust evolutionary algorithm framework called dual-stage robust evolutionary algorithm (DREA) for finding robust solutions.
Unlike other methods for evolutionary robust optimization, DREA utilizes points with good fitness values from the original problem (without considering the uncertainties) to aid in searching for robust optimal solutions.

DREA operates in two stages for searching robust solutions: the peak-detection stage and the robust solution-searching stage.
The primary aim of the peak-detection stage is to identify peaks in the fitness landscape of the original optimization problem.
This is achieved by employing an external archive that stores historical population and fitness values during this stage. Subsequently, in the robust solution-searching stage, the information gathered from the peak-detection stage is utilized to efficiently locate the robust optimal solution.
These two stages collectively enable DREA to efficiently attain the robust optimal solution for the optimization problem.
We conducted comprehensive experiments to verify the effectiveness of DREA.
It was compared with five popular algorithms across 18 test problems with diverse complex characteristics.
The experimental results demonstrate that DREA significantly outperforms the other five algorithms.
Additionally, we analyzed the search behavior of DREA, revealing that the peaks discovered in the first stage facilitate a quicker convergence towards the robust optimal solution in the second stage.
Furthermore, experiments analyzing the sensitivity of the parameter $N_p$ indicate that DREA with $N_p=3$ performs promisingly across both types of test problems.
Moreover, we evaluated the performance of DREA on higher-dimensional robust optimization problems.
The results indicate that when the test problems are extended to 100-$D$ and 200-$D$, DREA surpasses all five counterpart algorithms in comparison.

Compared with existing research in the literature, we innovated in several aspects to efficiently handle robust optimization problems.
Firstly, we directly utilize the optimal solutions of the original optimization problem, disregarding perturbations, in the search for robust optimal solutions.
This approach achieves a balance between solution optimality and robustness performance by separating the search process of optimal and robust optimal solutions.
Secondly, by leveraging historical solutions visited in both the decision space and objective space, we present an efficient peak-detection method to identify specific peaks.
This operation implies the idea of implicit averaging and saves the computational cost of locating the related peaks from avoiding extra explicit sampling.
Thirdly, we achieve seamless integration of optimality and robustness through the dual-stage operation, a novel framework compared to state-of-the-art optimization techniques for evolutionary robust optimization.

In the future, we plan to extend the proposed DREA to solve real-world robust optimization problems and address more complex optimization problems, such as constrained optimization problems, multi/many-objective optimization problems, and so on.

\ifCLASSOPTIONcaptionsoff
  \newpage
\fi

{\footnotesize\bibliography{reference/ref}
\bibliographystyle{ieeetr}}

\end{document}